\documentclass[10pt,twocolumn,letterpaper]{article}

\usepackage{wacv}
\usepackage{times}
\usepackage{epsfig}
\usepackage{graphicx}
\usepackage{amsmath}
\usepackage{amssymb}
\usepackage{graphicx,xparse,booktabs}
\ExplSyntaxOn
\NewDocumentEnvironment{places}{mm}
 {% #1 is the desired width, #2 is the number of photos per line
  \setlength{\tabcolsep}{0pt} % no space between rows
  \dim_set:Nn \l_places_width_dim
   {
    (#1-\ht\strutbox-\dp\strutbox-2pt)/(#2)
   }
  \begin{tabular}{r @{\hspace{2pt}} *{#2}{c}}
 }
 {
  \end{tabular}
 }

\NewDocumentCommand{\place}{mm}
 {% #1 is the name of the place, #2 is the comma separated list of images
  \seq_set_from_clist:Nn \l_places_images_in_seq { #2 }
  \seq_set_map:NNn \l_places_images_out_seq \l_places_images_in_seq { \places_set_image:n {##1} }
  \seq_put_left:Nn \l_places_images_out_seq
   {
    \begin{tabular}{c}\rotatebox[origin=c]{90}{\strut#1}\end{tabular}
   }
  \seq_use:Nn \l_places_images_out_seq { & } \\ \addlinespace
 }

\dim_new:N \l_places_width_dim
\seq_new:N \l_places_images_in_seq
\seq_new:N \l_places_images_out_seq

\cs_new_protected:Nn \places_set_image:n
 {
  \makebox[\l_places_width_dim]
   {
    \begin{tabular}{c}
    \includegraphics[
      width=\l_places_width_dim,
      height=\l_places_width_dim,
      keepaspectratio,
    ]{#1}
    \end{tabular}
   }
 }

\ExplSyntaxOff

\usepackage{amsmath}
\usepackage{amssymb}
\usepackage{booktabs}
\usepackage{graphics}
\usepackage{epstopdf}
\usepackage{multirow}
\usepackage{subcaption}
\usepackage{adjustbox}
\usepackage{color}

\usepackage{xcolor}

\usepackage{amsthm}

\usepackage{mathtools}
\DeclarePairedDelimiter{\norm}{\lVert}{\rVert}

\mathchardef\dash="2D

\wacvfinalcopy 
\def\wacvPaperID{203} % Enter the WACV Paper ID here

\wacvalgorithmstrack   % Uncomment this line if you are submitting to the Algorithms Track.
\ifwacvfinal
\usepackage[breaklinks=true,bookmarks=false]{hyperref}
\else
\usepackage[pagebackref=true,breaklinks=true,colorlinks,bookmarks=false]{hyperref}
\fi

\usepackage[capitalize]{cleveref}
\crefname{section}{Sec.}{Secs.}
\Crefname{section}{Section}{Sections}
\Crefname{table}{Table}{Tables}
\crefname{table}{Tab.}{Tabs.}

\begin{document}

%%%%%%%%% TITLE
\title{Evaluating Adversarial Robustness in the Spatial Frequency Domain}

\author{Keng-Hsin Liao, Chin-Yuan Yeh, Hsi-Wen Chen, Ming-Syan Chen\\
National Taiwan University\\
{\tt\small cyyeh@arbor.ee.ntu.edu.tw}
% For a paper whose authors are all at the same institution,
% omit the following lines up until the closing ``}''.
% Additional authors and addresses can be added with ``\and'',
% just like the second author.
% To save space, use either the email address or home page, not both
% \and
% Second Author\\
% Institution2\\
% First line of institution2 address\\
% {\tt\small secondauthor@i2.org}
}

\maketitle
\thispagestyle{empty}

%%%%%%%%% ABSTRACT
\begin{abstract}
    Convolutional Neural Networks (CNNs) have dominated the majority of computer vision tasks. However, CNNs' vulnerability to adversarial attacks has raised concerns about deploying these models to safety-critical applications. In contrast, the Human Visual System (HVS), which utilizes spatial frequency channels to process visual signals, is immune to adversarial attacks. As such, this paper presents an empirical study exploring the vulnerability of CNN models in the frequency domain. Specifically, we utilize the discrete cosine transform (DCT) to construct the \emph{Spatial-Frequency (SF)} layer to produce a block-wise frequency spectrum of an input image and formulate \emph{Spatial Frequency CNNs (SF-CNNs)} by replacing the initial feature extraction layers of widely-used CNN backbones with the SF layer. Through extensive experiments, we observe that SF-CNN models are more robust than their CNN counterparts under both white-box and black-box attacks. To further explain the robustness of SF-CNNs, we compare the SF layer with a trainable convolutional layer with identical kernel sizes using two mixing strategies to show that the lower frequency components contribute the most to the adversarial robustness of SF-CNNs. We believe our observations can guide the future design of robust CNN models.
\end{abstract}

% Convolutional Neural Networks (CNNs) have dominated the majority of computer vision tasks. However, CNNs' vulnerability to adversarial attacks has raised concerns about deploying these models to safety-critical applications. In contrast, the Human Visual System (HVS), which utilizes spatial frequency channels to process visual signals, is immune to adversarial attacks. As such, this paper presents an empirical study exploring the vulnerability of CNN models in the frequency domain. Specifically, we utilize the discrete cosine transform (DCT) to construct the Spatial-Frequency (SF) layer to produce a block-wise frequency spectrum of an input image and formulate Spatial Frequency CNNs (SF-CNNs) by replacing the initial feature extraction layers of widely-used CNN backbones with the SF layer. Through extensive experiments, we observe that SF-CNN models are more robust than their CNN counterparts under both white-box and black-box attacks. To further explain the robustness of SF-CNNs, we compare the SF layer with a trainable convolutional layer with identical kernel sizes using two mixing strategies to show that the lower frequency components contribute the most to the adversarial robustness of SF-CNNs. We believe our observations can guide the future design of robust CNN models.

%%%%%%%%% BODY TEXT
\section{Introduction}
    Convolutional neural networks (CNNs) have excelled at computer vision tasks, including image classification~\cite{he2016deep}, object detection~\cite{redmon2016you}, and image segmentation~\cite{liu2022convnet}. However, it has been demonstrated that these models are vulnerable to adversarial attacks, which apply imperceptible perturbations on input images to \emph{fool} targeted CNN models, \eg, to cause wrong classification results. This raises concerns about whether current CNN models should be deployed on safety-critical applications~\cite{goodfellow2018defense}. 
    One approach is to try to improve robustness by training CNN models with their adversarial examples, which is known as adversarial training~\cite{madry2018towards, athalye2018obfuscated, tramer2020adaptive,wong2020fast,wu2020adversarial}. Nevertheless, CNN models still could not function reliably when facing adversarial attacks. Croce \etal~\cite{croce2020robustbench} show that state-of-the-art defense strategies could not surpass $60\%$ accuracy on the ImageNet dataset. Other disadvantageous of adversarial training includes a reduction in accuracy on natural unaltered inputs~\cite{stutz2019disentangling,madry2018towards} and a significant increase in training cost~\cite{shafahi2019adversarial}.

    As an alternative route, researchers have focused on developing more robust architectures to defend against adversarial attacks~\cite{{vuyyuru2020biologically, dapello2020simulating}}. Because humans are not fooled by (and may not even notice) adversarial examples, several studies have been conducted to investigate the Human Visual System (HVS) to find possible mechanisms that can improve adversarial robustness. Vuyyuru~\etal~\cite{vuyyuru2020biologically}, for example, mimic human eye fixations with a non-uniform sampling of multiple receptive fields at different sizes and locations to strengthen the robustness of CNNs. Dapello~\etal~\cite{dapello2020simulating} design convolutional layers to simulate a primary visual cortex to obtain better robustness. 
    
    Yet, a fundamental understanding that the HVS utilizes spatial frequency information is absent in the above works. According to the spatial frequency theory~\cite{devalois1990spatial, martinez2003complex}, the HVS uses spatial frequency channels and is more sensitive to low-frequency changes than high-frequency changes~\cite{vuilleumier2003distinct, majaj2002role, sachs1971spatial}.\footnote{Spatial frequencies denote the periodic spectrum of visual signals across the image input, which can entail rich semantic information~\cite{xu2020learning}.} In contrast, recent efforts have demonstrated that CNN models may overly rely on human-imperceptible high-frequency patterns for downstream tasks~\cite{wang2020high, geirhos2018imagenettrained}, which also explains CNN's vulnerability towards adversarial examples~\cite{ilyas2019adversarialnotbugs, wang2020high}. On the other hand, several works deploy CNN models in the frequency domain~\cite{gueguen2018faster, ehrlich2019deep, xu2020learning}, which we denote as \emph{Spatial Frequency CNN (SF-CNN)}. Nevertheless, their works focus on task performance~\cite{qin2021fcanet} and computation efficiency~\cite{xu2020learning} by leveraging the widely used JPEG standard~\cite{wallace1992jpeg}. In contrast, the disparity between the HVS and the CNN designs prompts us to explore the potential impact of spatial frequency features on adversarial vulnerability. 

    We address these questions by presenting an empirical evaluation of the SF-CNN models compared with the normal CNN models. Specifically, we employ well-known CNN model structures (\ie, ResNet~\cite{he2016deep}, EfficientNet~\cite{tan2021efficientnetv2}, and DenseNet~\cite{huang2017densely}), but replace the first few convolutional layers with \emph{Spatial Frequency (SF)} layer, which retrieves a block-wise spatial frequency spectrum. By these models, we compare the adversarial robustness of a convolution-based feature extractor (the first few CNN layers) with a spatial frequency feature extractor.\footnote{In practice, we divide the input images into $8\times8$ pixel blocks, then utilize the Discrete Cosine Transform (DCT)~\cite{ahmed1974discrete} to extract horizontal and vertical spatial frequency components within each block. Our spatial frequency extraction process closely resembles the widely used JPEG compression standard~\cite{wallace1992jpeg}, which was also inspired by the use of spatial frequencies in the HVS.} Our experiments demonstrate that SF-CNNs outperform standard CNNs in terms of model robustness under both white-box and black-box attacks. For all evaluated datasets, SF-CNNs achieve \emph{higher} accuracy than CNN models with similar architectures under the standard white-box attack, (\ie, the Projected Gradient Descent attack~\cite{madry2018towards}). Furethermore, SF-CNNs are much more robust than CNNs when facing adversarial examples generated from other CNN and SF-CNN models (\ie, transfer attacks~\cite{papernot2016transferability}). Finally, we compare the SF layer to a trainable convolutional layer and find that SF-CNN models utilize low-frequency channels better than CNN models. In addition, with two mixing strategies, \ie, \emph{interpolation} and \emph{substitution}, we observe that the kernel weights of the first feature extraction layer contribute significantly to the robustness of a model.
    
    The following is a summary of contributions.
\begin{itemize}
    \item To the best of our knowledge, this is the first work that provides a thorough evaluation of the adversarial robustness of SF-CNNs.
    \item This paper investigates the performance of SF-CNNs under white-box and back-box attacks empirically. The experimental results show that using spatial frequency channels as a feature extractor, SF-CNNs outperform standard CNN models under all types of attacks.
    \item To further investigate the SF layer, we propose two mixing strategies between SF-CNNs and standard CNNs. The results show that the use of lower frequency components is related to the robustness of SF-CNNs.
\end{itemize}

\section{Related works}
\subsection{Adversarial attack and defense}
    The objective of an adversarial attack is to create adversarial examples that deceive the targeted model by introducing human-imperceptible perturbations to natural images~\cite{szegedy2013intriguing}. Based on the attacker's capabilities, adversarial attacks can be categorized into white-box or black-box attacks. Under the white-box setting, attackers have direct access to the model weights and may perform gradient calculations~\cite{goodfellow2014explaining, madry2018towards} or other optimization techniques~\cite{szegedy2013intriguing, carlini2017towards} to effectively obtain the precise adversarial patterns. As a standard first-order attack for evaluation of adversarial robustness~\cite{zhang2019theoretically, vuyyuru2020biologically}, the Projected Gradient Descent (PGD) attack~\cite{madry2018towards} generates adversarial examples using an iterative gradient descent operation. Under the black-box setting, the attackers may employ the aforementioned gradient schemes by querying the target model to estimate the gradient via zero-order optimizations~\cite{tu2019autozoom}. Otherwise, they may construct \emph{surrogate models} to conduct \emph{transfer attacks}, in which adversarial examples generated from a surrogate model are applied to the targeted model. The empirical evidence for standard CNNs' vulnerability to transfer attacks is well documented~\cite{papernot2016transferability, liu2016delving}. Demontis~\etal~\cite{demontis2019adversarial} show, in particular, that the complexity of the surrogate model and the target model influences the transferability of adversarial examples. Inkawhich~\etal~\cite{inkawhich2019feature} demonstrate that directly perturbing in the feature space of the hidden layers yields adversarial examples that are highly transferable.  

    To defend against adversarial attacks, many researchers propose different adversarial training strategies that retrain the target model on its adversarial examples to achieve better robustness~\cite{goodfellow2014explaining, madry2018towards}. Nonetheless, the approach is unsatisfactory due to the high computation cost~\cite{tsipras2018robustness} and the trade-off between standard accuracy and robustness~\cite{zhang2019theoretically}.
    % , and the preferential robustness against encountered attack methods~\cite{zhang2018the}. 
    Thus, some researchers have also sought alternative mechanisms that may help bring the models closer to human-level robustness~\cite{vuyyuru2020biologically}. For instance, Kang~\etal~\cite{kang2021stable} leverage neural Ordinary Differential Equations to maintain a stable model response towards adversarial perturbations. Dong~\etal~\cite{dong2021towards} achieve robustness with AdderNets, which replaces the multiplications in a convolutional layer with additions. However, the aforementioned works require substantial modifications to the standard CNN design. In contrast, we investigate the possibility of achieving adversarial robustness simply by utilizing the spatial frequencies as the initial feature for CNN models.
    
\subsection{Learning in the frequency domain}
\label{subsec:FL}
    Since condensed signals in the spatial frequency domain provide much information for computer vision tasks, several recent works incorporate the spatial frequency extraction process in CNNs for task performance~\cite{xu2020learning, qin2021fcanet}, computation efficiency~\cite{gueguen2018faster, ehrlich2019deep}, or model compression~\cite{lavin2016fast,chen2016compressing}. Gueguen~\etal~\cite{gueguen2018faster} and Ehrlich~\etal~\cite{ehrlich2019deep} modify the CNN models to encode the compressed JPEG code~\cite{wallace1992jpeg} for a faster operation speed. In addition, Qin~\etal~\cite{qin2021fcanet} include an attention module on the frequency channels to improve image classification accuracy. Motivated by these initial attempts, subsequent research extended frequency learning to a variety of tasks, including domain generalization~\cite{huang2021fsdr}, adaptation~\cite{huang2021rda}, and instance segmentation~\cite{shen2021dct}. Furthermore, the frequency spectrum has also been used to investigate standard CNNs. Wang~\etal~\cite{wang2020high} argue that CNNs rely on high-frequency components since CNNs achieve higher accuracy on high-frequency reconstructions of test images. They also empirically showed that smoothing the kernel weights can often provide adversarial robustness despite causing substantial reductions in task accuracy as well. Nevertheless, with these advancements, the adversarial robustness of CNN models trained in the spatial frequency domain has not been assessed.

\section{Methodology}
\label{sec:3_method}
    In this study, we examine the adversarial robustness of CNNs that operate in the spatial frequency domain as opposed to the conventional CNNs that operate with image pixels. For clarity, we refer to the former as \emph{Spatial Frequency Convolutional Neural Networks (SF-CNN)}. SF-CNNs are formed by substituting the initial layers of standard CNN models with the spatial frequency (SF) layer that employs a spatial frequency extraction procedure. We will first describe the mechanism of the SF layer, then present the architecture of the SF-CNN models.

\subsection{Spatial frequency layer}
\label{sec3_dct}
    Here, we introduce the spatial frequency (SF) layer~\cite{gueguen2018faster, xu2020learning}, which employs a spatial frequency extraction procedure similar to the JPEG standard~\cite{wallace1992jpeg}. The SF layer transforms an $H\times W\times 3$ color image into an $H/8\times W/8\times 192$ output where the input image is split into $8\times8$ pixel blocks and each color channel of each pixel block is separately filtered by the Discrete Cosine Transform (DCT). DCT extracts the spatial frequencies along both horizontal and vertical directions.  With the $8\times8$ pixel block represented by the matrix $\mathbf{B}$, DCT obtains an $8\times8$ frequency matrix $\mathbf{D}$ as follows.
    \begin{align}\small
        \label{eqn:DCT}
        \mathbf{D}_{u,v}&= \sum _{x=0}^7\sum _{y=0}^7\mathbf{B}_{x,y}\mathbf{K}_{x,y}^{u,v}\\
        \label{eqn:DCT_kernel}
        \mathbf{K}_{x,y}^{u,v} &= 
        \frac{c_u c_v}{4}\cos \left[{\frac {(2x+1)u\pi }{16}}\right]
        \cos \left[{\frac {(2y+1)v\pi }{16}}\right]
    \end{align}
    where $u$ and $v$ denote the horizontal and vertical spatial frequencies, respectively, and $c_u$ is the normalizing factor equaling to $\frac{1}{\sqrt{2}}$ if the index $u = 0$, otherwise, $c_u = 1$. Afterwards, the frequency matrix $\mathbf{D}$ is flattened into a $64\dash$element vector by a zigzag pattern scan from low frequency to high frequency, and the flattened vectors of the three color channels of the same pixel block are concatenated together. Thus, the image is transformed from $(H, W, 3)$ to $(H/8, W/8, 3\times 64)$ where $H$ and $W$ are the image height and the width, respectively.
    
    The SF layer can be considered a particular case of applying convolutional layers with fixed weights $\mathbf{K}^{u,v}_{x,y}$. Specifically, the above block-wise DCT process can be replicated by an $8\times8$ convolutional layer with stride $8$ using \cref{eqn:DCT_kernel} as the kernel weights. While the SF layer shares a similar function space as convolutional layers, our subsequent experiments demonstrate that SF-CNNs learn more robust parameters than standard CNNs because the SF layer can retain more low-frequency patterns (detailed in Sec.~\ref{sec:5_qual}).

% Remark: 

\begin{figure}[t]
  \vspace{.2em}
  \includegraphics[width=\linewidth]{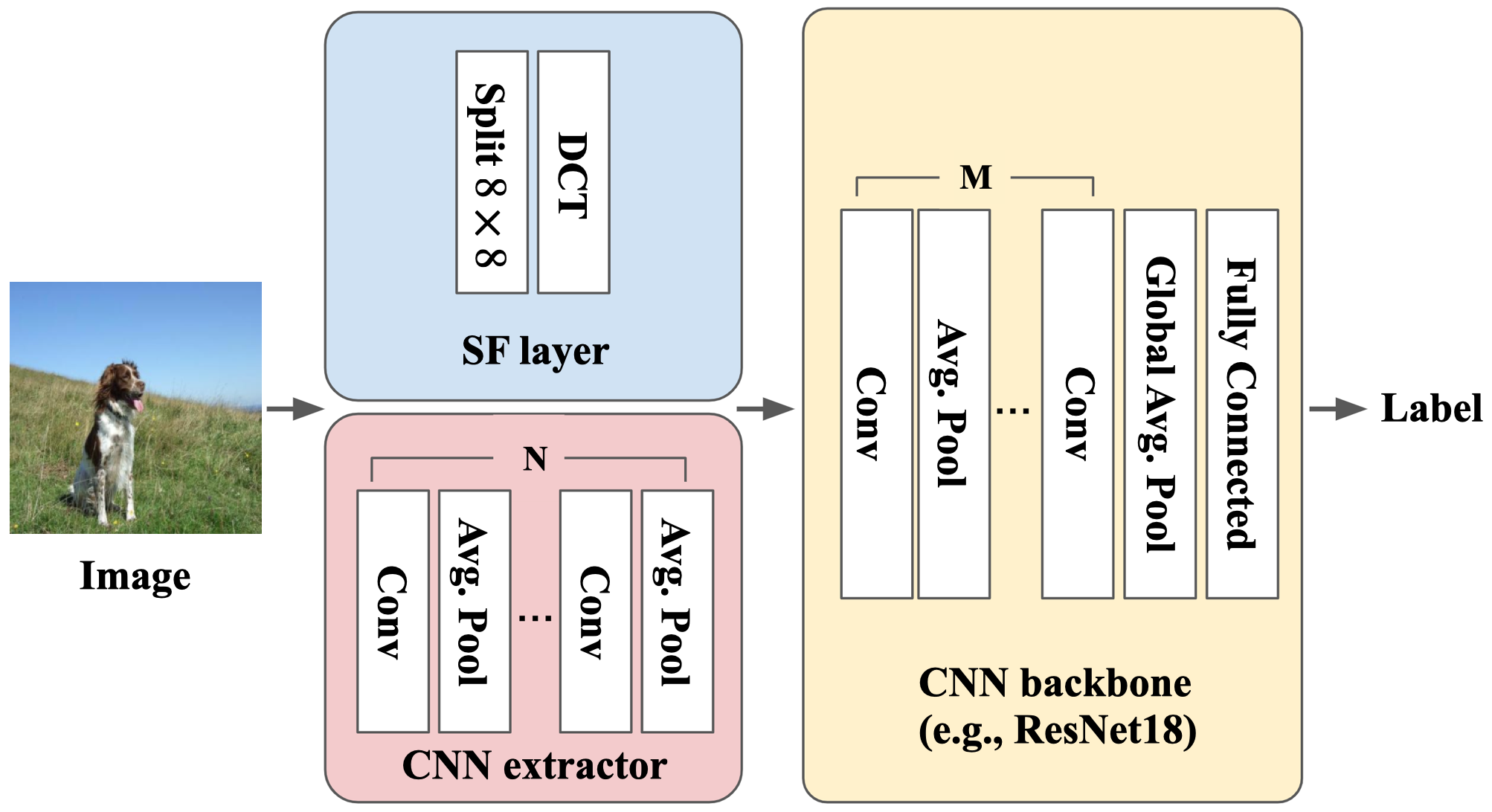}
  \vspace{.2em}
  \caption{The comparison between SF-CNNs and standard CNNs. Standard CNNs utilize $N$ trainable convolutional layers in the feature extractor and backbone classifier. SF-CNNs replace the feature extractor with the SF layer which conducts the block-wise DCT to extract the spatial frequencies as the initial image features but follows the same $M$-layer backbone architecture for classification.}
  \label{fig:sfcnn_config}
\end{figure}

\subsection{Spatial frequency CNN}
\label{subsec:3_fcnn}

    % CNNs excel in computer vision tasks by scanning and discovering local patterns over the whole image \cite{liu2022convnet}. 
    To create the SF-CNNs, we modify widely-used CNN architectures by replacing the first few layers of each experimented model with the SF layer. In particular, by modifying \textit{i) ResNet18}\cite{he2016deep}, \textit{ii) EfficientNet}\cite{tan2021efficientnetv2} and \textit{iii) DenseNet}\cite{huang2017densely}, we create SF-ResNet18, SF-DenseNet, and SF-EfficientNet, which are employed in the following adversarial robustness evaluations. In addition, SF-VGG11 is created from VGG11~\cite{simonyan2014very} for a basic CNN model to be utilized in the transfer attack experiments. The modification guidelines are introduced as follows. We consider the SF layer to be a feature extractor with $192$ different filters, corresponding to the different (block-wise) spatial frequencies of the three color spaces. \cref{fig:sfcnn_config} compares the SF-CNN with conventional CNN. Since most CNN model designs have monotonically increasing channel numbers for deeper layers~\cite{heo2021rethinking}, we replace the first few convolutional layers with less than $192$ channels from the original CNN model with the SF layer. The subsequent layer would receive a $192$ channel input instead of the original number. Due to space constraints, details of each experimented model architecture are provided in \cref{apx:A1_architect} .

\section{Adversarial robustness of SF-CNN}
\label{sec:4_Experiments}

    In this section, we examine the performance and the robustness of SF-CNNs under both white-box and black-box attacks. While black-box attacks can be categorized into transfer attacks~\cite{demontis2019adversarial} or query-based attacks~\cite{tu2019autozoom}, the latter follows the PGD framework but estimates the gradient with zero-order queries. Thus, we focus on the transfer attack to evaluate different types of adversarial threats. We utilize \textit{i) imagenette}~\cite{imagenette} which selects $10$ distinct and easily identifiable classes out of the total $1000$ classes, as well as \textit{ii) CIFAR10}~\cite{krizhevsky2009learning} with $10$ classes of tiny images and \textit{iii) Flower102}~\cite{nilsback2008automated} with $102$ flower categories of large images to assert the generality of our observations.  We split each dataset by $80\%$, $10\%$, and $10\%$, for training, validation, and testing, respectively.\footnote{To ensure integral division, the image datasets are reshaped to multiples of $8$, where the ImageNette, CIFAR10, and Flower102 images are resized to ($256 \times 256$), ($96 \times 96$) and ($320 \times 320$), respectively.} By default, we perform $100\dash$step PGD attacks for thorough evaluations of model robustness in terms of accuracy. All model parameters are initialized with the Glorot initialization~\cite{glorot2010understanding} and trained by the Adam optimizer with the default hyperparameter settings~\cite{kingma2014adam}. The training and evaluation procedures are implemented on an HP DL580 Gen 9 server with four 3.0 GHz Intel CPUs, 1TB RAM, and 8 Titan RTX 3080ti GPUs.

\subsection{Robustness against white-box attacks}
\label{subsec:4-1white}
\begin{table*}[ht]\small
    \centering
    \begin{tabular}{lc|cc c|cc c|cc}
    \toprule
    &\multicolumn{3}{c}{ImageNette} & \multicolumn{3}{c}{CIFAR10} & \multicolumn{3}{c}{Flower102}\\
    \cmidrule(lr){2-4}\cmidrule(lr){5-7}\cmidrule(lr){8-10}
    model
    & Acc. &  $\epsilon =0.003$ & $\epsilon =0.01$ 
    & Acc. &  $\epsilon =0.003$ & $\epsilon =0.01$ 
    & Acc.&  $\epsilon =0.003$ & $\epsilon =0.01$ \\
    \midrule
    SF-VGG11 & $78.8\%$ & $63.6\%$ & $33.8\%$ & $78.6\%$ & $62.2\%$ & $42.0\%$ & $70.1\%$ & $54.7\%$ & $24.5\%$\\
    VGG11 & $78.6\%$ & $60.2\%$ & $21.0\%$ & $79.0\%$ & $61.4\%$ & $27.8\%$ & $70.9\%$ & $0.0\%$ & $0.0\%$\\
    % (improvement) && ($+3.4\%$) & ($+12.8\%$)  && ($+0.8\%$) & ($+14.2\%$)  && ($+54.7\%$) & ($+24.5\%$) \\
    \cmidrule(lr){2-4}\cmidrule(lr){5-7}\cmidrule(lr){8-10}
    SF-ResNet18 & $80.8\%$ & $67.2\%$ & $31.4\%$ & $78.0\%$ & $62.7\%$ & $26.4\%$ & $75.1\%$ & $61.7\%$ & $29.7\%$\\
    ResNet18 & $80.2\%$ & $57.6\%$ & $14.4\%$ & $78.0\%$ & $60.8\%$ & $24.1\%$ & $74.0\%$ & $58.9\%$ & $16.4\%$\\
    % (improvement) && ($+9.6\%$) & (+$17.0\%$)  && ($+1.9\%$) & ($+2.3\%$)  && ($+2.8\%$) & ($+13.3\%$) \\
    \cmidrule(lr){2-4}\cmidrule(lr){5-7}\cmidrule(lr){8-10}
    SF-EfficientNet & $74.0\%$ & $58.2\%$ & $21.6\%$ & $77.0\%$ & $52.9\%$ & $8.6\%$ & $80.1\%$ & $55.2\%$ & $10.0\%$\\
    EfficientNet & $74.0\%$ & $45.4\%$ & $2.8\%$ & $79.8\%$ & $26.8\%$ & $0.2\%$ & $81.2\%$ & $43.8\%$ & $3.4\%$\\
    % (improvement) && ($+12.8\%$) & ($+18.8\%$)  && ($+26.1\%$) & ($+8.4\%$)  && ($+11.4\%$) & ($+6.6\%$) \\
    \cmidrule(lr){2-4}\cmidrule(lr){5-7}\cmidrule(lr){8-10}
    SF-DenseNet & $78.2\%$ & $61.2\%$ & $28.2\%$ & $77.1\%$ & $58.4\%$ & $18.5\%$ & $73.0\%$ & $60.8\%$ & $32.6\%$\\
    DenseNet & $78.2\%$ & $60.2\%$ & $25.6\%$ & $79.2\%$ & $19.6\%$ & $0.8\%$ & $71.4\%$ & $53.9\%$ & $14.8\%$\\
    % (improvement) && ($+1.0\%$) & ($+2.6\%$)  && ($+38.8\%$) & ($+17.7\%$)  && ($+6.9\%$) & ($+17.8\%$) \\

    \bottomrule 
    \end{tabular}
    \caption{Model performances under PGD attack in the pixel domain. ``Acc.'' denotes the test accuracy on clean images.}
    \label{tab:exp_1_norm_bound_asr}
\end{table*}

    To analyze the robustness of SF-CNNs, we follow the standard adversarial robustness evaluation protocol to utilize PGD attacks~\cite{zhang2019theoretically}, which generates the adversarial examples in an iterative gradient optimization process, in both pixel domain and frequency domain.
    
    \subsubsection{Perturbation in the pixel domain}
    In the following, we first follow a conventional attack schemes~\cite{goodfellow2014explaining} to perturb the original images on its pixel values. Given the target classifier model $C$, the norm-bounded attack aims to find an adversarial example $x^*$ for a given test image $x$ such that,
    \begin{equation}\small
        \label{eqn:background_norm_bound}
            C(x^*) \neq C(x), 
            \,\, s.t. \,\, 
            x^*\in[0,1]^N \, \cap \,\norm{x^*-x}_\infty \leq 
       \epsilon,
    \end{equation}
    where $[0, 1]^N$ indicates that each of the $N$ pixels is in the legitimate range of values. Note that $\norm{x^*-x}_\infty \leq \epsilon$ restricts the modification to the test image to be human-imperceptible. We denote the union of the two limits as the adversarial limit $\Omega_\epsilon \equiv[0,1]^N \, \cap \,\norm{x^*-x}_\infty \leq \epsilon$. With full knowledge of the model weights, the PGD attack generates adversarial examples with the following iterative scheme:
    \begin{equation}\small
    \label{eqn:attack_pixel}
        x^*_{t+1} = 
        \Pi_{\Omega_\epsilon}(x^*_t + \eta \,\text{sign}(\nabla\mathcal{L}(C, x^*_t, y))),
    \end{equation}
    where $t$ is the count of iteration with $x^*_0$ being the original test image, $\Pi_{\Omega_\epsilon}$ is the projection operation that projects the adversarial example back to the adversarial limit, $\eta$ is the step size, $\nabla\mathcal{L}$ is the gradient of the relevant attack loss function, and $y$ is the correct classification $C(x^*_0) = y$.
    
    Following~\cite{vuyyuru2020biologically}, we set $\epsilon = 0.003$ and $\epsilon = 0.01$ with $\eta=0.001$ and $0.003$, respectively. \Cref{tab:exp_1_norm_bound_asr} shows that all SF-CNN models trained in all datasets are more robust than their CNN counterparts. In particular, SF-DenseNet trained in the CIFAR10 dataset presents a drastic $38.8\%$ increase in accuracy under $\epsilon=0.003$ attack without utilizing any adversarial defense methods. The result supports the idea of improving the adversarial robustness of CNNs by mimicking the HVS to utilize spatial frequency channels. Moreover, since the SF-CNN models significantly reduce the parameter size by removing several layers from the CNN model (\cref{fig:sfcnn_config}) but maintains similar test accuracy, we conclude that spatial frequency features provide sufficient semantic information for the model to learn.

\subsubsection{Perturbation in the frequency domain}

% \begin{table}[ht]\small
%     \centering
%     \begin{tabular}{ll|ccc}
%     \toprule
%     dataset & Acc. & $\epsilon_f=0.003$ & $\epsilon_f=0.01$\\
%     \midrule
%     ImageNette & $80.8\%$ & $5.6\%$ & $0.0\%$\\
%     CIFAR10 & $78.0\%$ & $17.5\%$ & $0.1\%$\\
%     Flower102 & $75.1\%$ & $21.5\%$ & $0.0\%$\\
%     \bottomrule 
%     \end{tabular}
%     \caption{Attack on frequency domain of SF-ResNet18 with three different datasets. }
%     \label{tab:attack_frequency}
% \end{table}

\begin{table}[ht]\small
    \centering
    \resizebox{\linewidth}{!}{
    
    \begin{tabular}{lcccccc}
    \toprule
    &\multicolumn{2}{c}{ImageNette} & \multicolumn{2}{c}{CIFAR10} & \multicolumn{2}{c}{Flower102}\\
    \cmidrule(lr){2-3}\cmidrule(lr){4-5}\cmidrule(lr){6-7}
    model $\backslash\,\epsilon_f$
    & $0.003$ & $0.01$ 
    & $0.003$ & $0.01$ 
    & $0.003$ & $0.01$ \\
    \midrule
    SF-VGG11 & $1.0\%$ & $0.0\%$ & $11.5\%$ & $0.0\%$ & $6.5\%$ & $3.4\%$\\
    VGG11 & $0.0\%$ & $0.0\%$  & $0.0\%$ & $0.0\%$ & $0.0\%$ & $0.0\%$\\
    \cmidrule(lr){2-3}\cmidrule(lr){4-5}\cmidrule(lr){6-7}
    SF-ResNet18 & $5.6\%$ & $0.0\%$ & $17.5\%$ & $0.1\%$ & $21.5\%$ & $0.0\%$\\
    ResNet18 & $0.0\%$ & $0.0\%$ & $0.0\%$ & $0.0\%$ & $0.0\%$ & $0.0\%$\\
    \cmidrule(lr){2-3}\cmidrule(lr){4-5}\cmidrule(lr){6-7}
    SF-EfficientNet & $8.8\%$ & $0.2\%$ & $1.5\%$ & $0.0\%$ & $29.2\%$ & $0.1\%$\\
    EfficientNet & $0.0\%$ & $0.0\%$ & $0.0\%$ & $0.0\%$ & $0.1\%$ & $0.0\%$\\
    \cmidrule(lr){2-3}\cmidrule(lr){4-5}\cmidrule(lr){6-7}
    SF-DenseNet & $5.4\%$ & $0.2\%$ & $0.0\%$ & $0.0\%$ & $49.5\%$ & $5.6\%$\\
    DenseNet & $0.6\%$ & $0.0\%$ & $0.0\%$ & $0.0\%$ & $0.0\%$ & $0.0\%$\\
    
    \bottomrule 
    \end{tabular}
    }
    \caption{Model performances under PGD attack in the frequency domain with $\epsilon_f=0.003$ or $0.01$.}
    \label{tab:attack_frequency}
\end{table}

\begin{figure}[t]
    \centering
    \begin{subfigure}[t]{.49\linewidth}
        \centering
        \includegraphics[width=.49\linewidth]{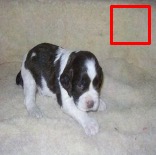}
        \includegraphics[width=.49\linewidth]{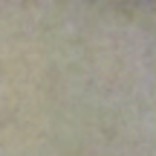}
        \includegraphics[width=.49\linewidth]{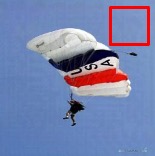}
        \includegraphics[width=.49\linewidth]{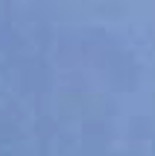}
        \includegraphics[width=.49\linewidth]{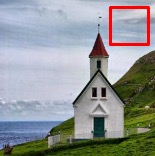}
        \includegraphics[width=.49\linewidth]{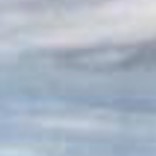}
        \caption{pixel domain}
    \end{subfigure}
    \begin{subfigure}[t]{.49\linewidth}
        \centering
        \includegraphics[width=.49\linewidth]{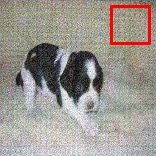}
        \includegraphics[width=.49\linewidth]{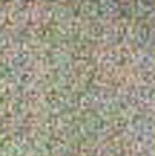}
        \includegraphics[width=.49\linewidth]{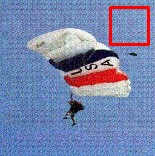}
        \includegraphics[width=.49\linewidth]{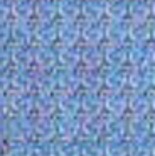}
        \includegraphics[width=.49\linewidth]{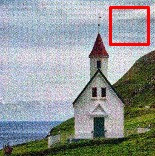}
        \includegraphics[width=.49\linewidth]{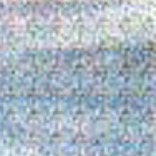}
        \caption{frequency domain}
    \end{subfigure}
    \caption{Examples of adversarial attacks in the pixel or the frequency domain with $\epsilon$ and $\epsilon_f$ set to $0.01$. Perturbation on the frequency domain leaves a more visible adversarial noise with a periodic pattern.}
    \label{fig:qualitative_evaluation}
    \end{figure}
    
    Next, we probe the SF-CNNs by perturbing the spatial frequencies $f$, \ie, the output of the SF layer, directly, as:
    \begin{equation}\small
        \label{eqn:attack_frequenct}
            C(x^*) \neq C(x), 
            \,\, s.t. \,\, 
            x^*\in[0,1]^N \, \cap \,\norm{f^*-f}_\infty \leq 
       \epsilon_f,
    \end{equation}
    where $f^*$ denotes the block-wise spatial frequencies of $x^*$. Similar to attacking in the pixel domain, we limit the perturbation by an $\epsilon_f$, and utilize the PGD attack to generate adversarial examples.
    
    As shown in \Cref{tab:attack_frequency}, we report that although SF-CNNs perform better than CNNs in several combinations, attacking in the frequency domain are successful in most cases where the accuracy of SF-CNNs and CNNs are both reduced to $0\%$. Particularly, the accuracy of CNNs are overall lower than $0.6\%$ while the accuracy of SF-CNNs are less than $49.5\%$ with $\epsilon_f = 0.003$ and less than $8.8\%$ with $\epsilon_f = 0.01$. However, we report that the $\epsilon_f\dash$bound corresponds to pixel differences that greatly exceed a corresponding $\epsilon\dash$bound. For instance, perturbing within $\epsilon_f =0.003$ of the frequency range results in a maximum of $0.122$ change in the pixel values. \cref{fig:qualitative_evaluation} also shows that such attacks leave distinct patterns that are visually observable in the adversarial examples. 
    
    By inspecting the frequency value distribution, we also find that attacking in the frequency domain create a significant shift in the spatial frequency value distribution, such that adversarial examples can be easily detected with a $100\%$ detection rate. In particular, while the distribution of frequency components forms a peak at the zero value, adversarial examples created by the PGD attack in the frequency domain shows two peaks corresponding to the $\epsilon_f$ limit. Details of a basic detection method based on distribution of frequency values are reported in \cref{apx:detection}. 
\subsection{Robustness against transfer attack}
\label{subsec:4-2transfer}
\begin{table}[t]\small
    \centering
    \resizebox{\linewidth}{!}{
    \begin{tabular}{lc|ccc}
    \toprule
    model & Acc. & $\epsilon =0.1$ &     $\epsilon =0.2$ & $\epsilon =0.3$ \\
    \midrule
    SF-VGG11 & $78.8\%$ & $73.8\%$ & $57.8\%$ & $44.2\%$\\
    VGG11 & $78.6\%$ & $0.0\%$ & $0.0\%$ & $0.0\%$\\
    \hline
    SF-ResNet18 &$80.8\%$ & $72.2\%$ & $49.0\%$ & $31.2\%$\\
    ResNet18 &$80.2\%$ & $52.2\%$ & $26.6\%$ & $16.0\%$\\
    \hline
    SF-EfficientNet &$74.2\%$ & $65.8\%$ & $42.2\%$ & $19.4\%$\\
    EfficientNet &$74.0\%$ & $63.6\%$ & $39.0\%$ & $16.8\%$\\
    \hline
    SF-DenseNet &$78.2\%$ & $72.8\%$ & $60.2\%$ & $40.8\%$\\
    DenseNet &$78.2\%$ & $60.8\%$ & $24.8\%$ & $11.8\%$\\
    \bottomrule 
    \end{tabular}
    }
    \caption{Model performances under transfer attack from VGG11. SF-CNNs are much more robust than CNNs.}
    \label{tab:exp_3_transfer}
\end{table}

\begin{table}[t]\small
    \centering
    \resizebox{\linewidth}{!}{
    \begin{tabular}{lc|ccc}
    \toprule
    model & Acc. & $\epsilon =0.1$ &     $\epsilon =0.2$ & $\epsilon =0.3$ \\
    \midrule
    SF-VGG11 & $80.2\%$ & $0.0\%$ & $0.0\%$ & $0.0\%$\\
    VGG11 & $78.6\%$ & $61\%$ & $25.8\%$ & $17\%$\\
    \hline
    SF-ResNet18 & $80.8\%$ & $73.8\%$ & $58.8\%$ & $38.2\%$ \\
    ResNet18 & $80.2\%$ & $69.2\%$ & $37\%$ & $14.6\%$\\
    \hline
    SF-EfficientNet & $74.2\%$ & $64.6\%$ & $61.4\%$ & $56.2\%$\\
    EfficientNet & $74.0\%$ & $72.8\%$ & $59\%$ & $41.2\%$\\
    \hline
    SF-DenseNet & $78.2\%$ & $72.0\%$ & $66.0\%$ & $58.4\%$\\
    DenseNet & $78.2\%$ & $68.6\%$ & $35.6\%$ & $17.4\%$\\
    \bottomrule 
    \end{tabular}
    }
    \caption{Model performances under transfer attack from SF-VGG11. Remarkably, SF-CNNs are more robust.}
    \label{tab:exp_3_transfer_fmodel}
\end{table}
\begin{figure}[t]
    \centering
    \begin{subfigure}[t]{.24\linewidth}
        \centering
        \includegraphics[width=1\linewidth]{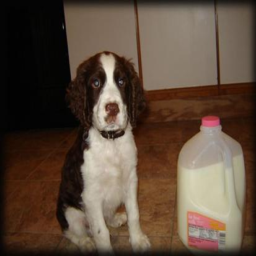}
        \includegraphics[width=1\linewidth]{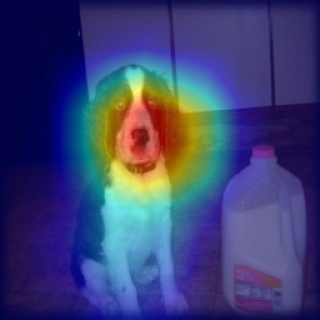}
        \includegraphics[width=1\linewidth]{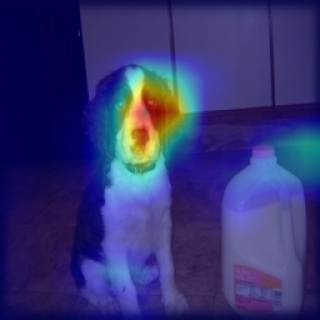}
        \caption*{$\epsilon = 0$}
    \end{subfigure}
    \begin{subfigure}[t]{.24\linewidth}
        \centering
        \includegraphics[width=1\linewidth]{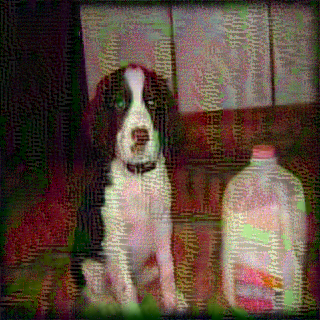}
        \includegraphics[width=1\linewidth]{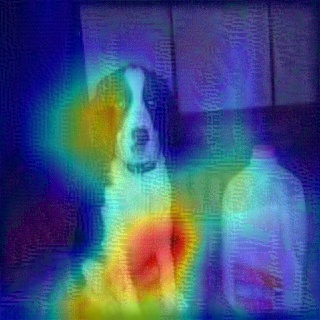}
        \includegraphics[width=1\linewidth]{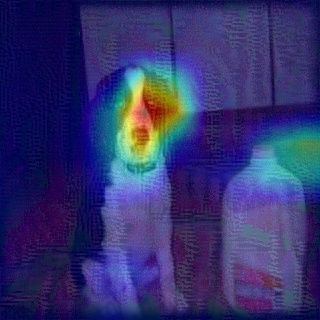}
        \caption*{$\epsilon = 0.1$}
    \end{subfigure}
    \begin{subfigure}[t]{.24\linewidth}
        \centering
        \includegraphics[width=1\linewidth]{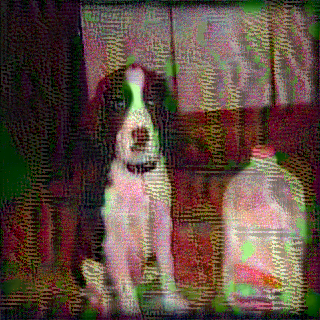}
        \includegraphics[width=1\linewidth]{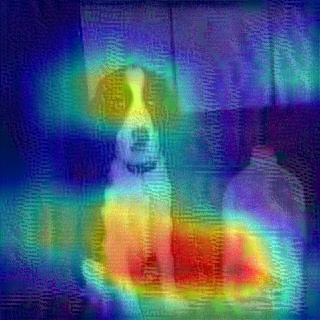}
        \includegraphics[width=1\linewidth]{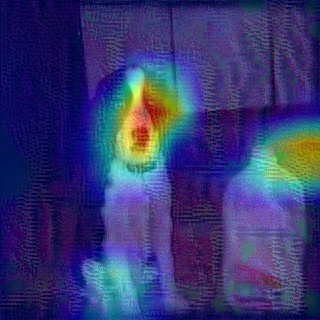}
        \caption*{$\epsilon = 0.2$}
    \end{subfigure}
    \begin{subfigure}[t]{.24\linewidth}
        \centering
        \includegraphics[width=1\linewidth]{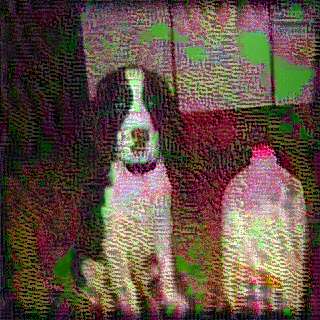}
        \includegraphics[width=1\linewidth]{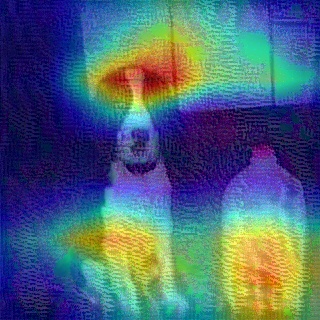}
        \includegraphics[width=1\linewidth]{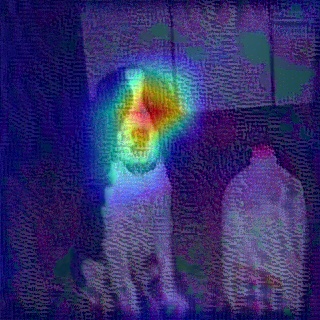}
        \caption*{$\epsilon = 0.3$}
    \end{subfigure}
    \caption{Grad-Cam visualizations of transfer attacks from VGG11. Top, middle, and bottom rows present the images, Grad-Cam visualization of ResNet18, and that of SF-ResNet18, respectively.}
    \label{fig:grad_cam}
\end{figure}

\begin{figure}[t]
    \centering
    \begin{subfigure}[t]{.24\linewidth}
        \centering
        \includegraphics[width=1\linewidth]{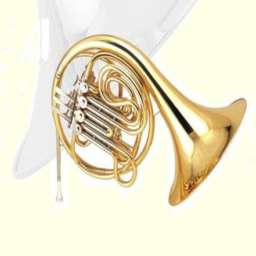}
        \includegraphics[width=1\linewidth]{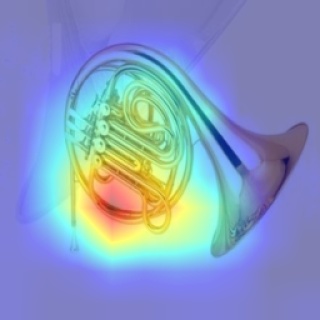}
        \includegraphics[width=1\linewidth]{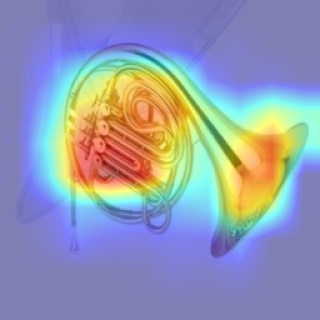}
        \caption*{$\epsilon = 0$}
    \end{subfigure}
    \begin{subfigure}[t]{.24\linewidth}
        \centering
        \includegraphics[width=1\linewidth]{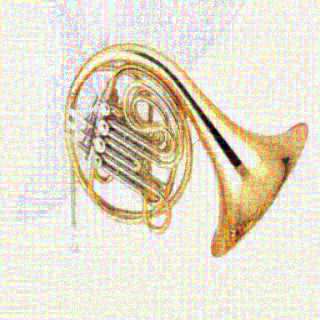}
        \includegraphics[width=1\linewidth]{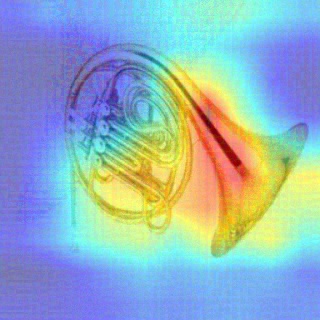}
        \includegraphics[width=1\linewidth]{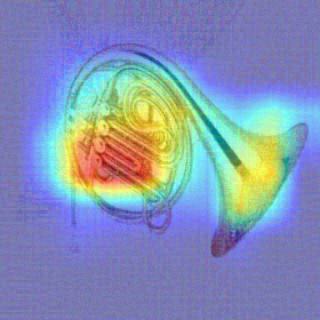}
        \caption*{$\epsilon = 0.1$}
    \end{subfigure}
    \begin{subfigure}[t]{.24\linewidth}
        \centering
        \includegraphics[width=1\linewidth]{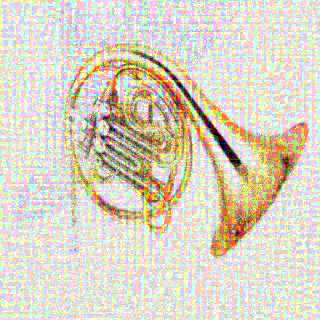}
        \includegraphics[width=1\linewidth]{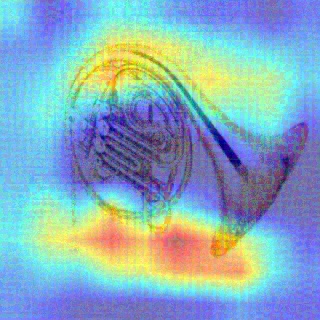}
        \includegraphics[width=1\linewidth]{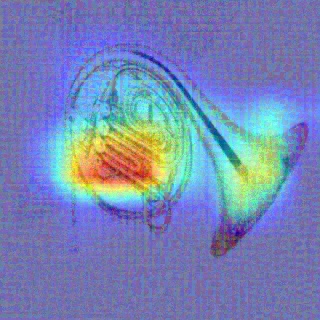}
        \caption*{$\epsilon = 0.2$}
    \end{subfigure}
    \begin{subfigure}[t]{.24\linewidth}
        \centering
        \includegraphics[width=1\linewidth]{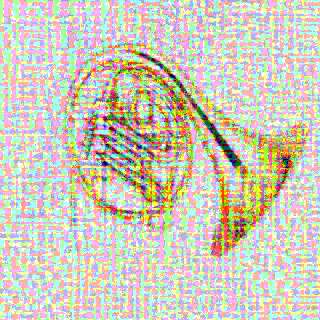}
        \includegraphics[width=1\linewidth]{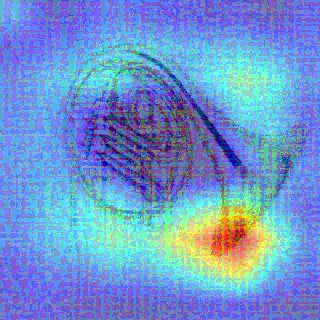}
        \includegraphics[width=1\linewidth]{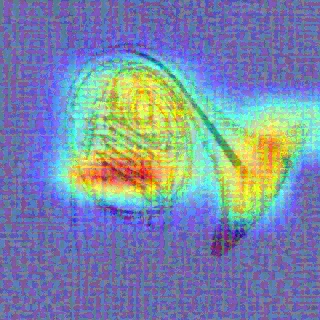}
        \caption*{$\epsilon = 0.3$}
    \end{subfigure}
    \caption{Grad-Cam visualizations of transfer attacks from SF-VGG11. Top, middle, and bottom rows present the images, Grad-Cam visualization of ResNet18, and that of SF-ResNet18, respectively.}
    \label{fig:grad_cam_sf}
\end{figure}

    Transferability of adversarial examples among CNN models is a well-known issue~\cite{goodfellow2014explaining}, leading to the possibility of \emph{transfer attacks}~\cite{demontis2019adversarial}. In the following, we compare the robustness of SF-CNN and CNN models to adversarial examples generated by a VGG11 model or a SF-VGG11 model using transfer attacks. In particular, the compared models including VGG11 and SF-VGG11 are all trained in the imagenette dataset, and the adversarial examples are created with PGD attacks.
    
    \subsubsection{Transfer attacks from VGG11}

    As shown in \Cref{tab:exp_3_transfer}, the model accuracy under transfer attacks from VGG11 is significantly \emph{higher} for all SF-CNN models compared to their CNN model counterparts. In particular, VGG11 models have $0\%$ accuracy since the adversarial examples are generated from the same models. However, transferring the same adversarial examples to SF-VGG11 remarkably find $73.8\%$, $57.8\%$ and $44.2\%$ under $\epsilon=0.1, 0.2$ and $0.3$, respectively. SF-ResNet18, SF-EfficientNet, and SF-DenseNet also on average enjoy $42.6\%$, $6.3\%$, and $56.9\%$ better accuracy under transfer attack, compared to ResNet18, EfficientNet, and DenseNet, respectively. A potential explanation can be found by inspecting the difference between the frequency domain and the image pixel domain. In particular, the basis of the frequency domain and the original image pixel domain are nearly orthogonal since the basis of the frequency domain corresponds to cosine values of different frequencies spanning the $8$ pixels in \cref{eqn:DCT_kernel}, whereas the basis of the pixel domain corresponds to each pixel. Note that we utilize $\epsilon$ values larger than the usual $0.03$~\cite{wu2020adversarial} to thoroughly compare SF-CNNs' and CNNs' resistance towards transfer attacks.

    By comparing Grad-CAM visualizations~\cite{selvaraju2017grad} of the model attention of SF-ResNet18 and that of ResNet18 on the adversarial examples, we find that SF-CNNs indeed learn a different set of features by operating in the spatial frequency domain. As shown in \cref{fig:grad_cam}, initially, both SF-ResNet18 and ResNet18 focus on the same semantic regions to correctly classify the clean test image. As the adversarial perturbation increases along with the $\epsilon$ value, the adversarial pattern begins to divert ResNet18's attention towards other regions, such that it could not focus on the correct regions. However, despite the intense adversarial noises, SF-ResNet18's attention maintains roughly at the same place and could still correctly classify the adversarial example. Such a result indicates that adversarial patterns that can fool CNNs are mostly neglected by SF-CNNs, allowing SF-CNNs to be robust against transfer attacks.\footnote{Grad-CAM visualizations of more models for both transfer attacks from VGG11 and SF-VGG11 are provided in \cref{apx:grad_cam}.}
    
\subsubsection{Transfer attacks from SF-VGG11}
    \Cref{tab:exp_3_transfer_fmodel} presents the results of the transfer attacks from SF-VGG11. Remarkably, all SF-CNN models (except SF-VGG11 itself) perform better than their CNN counterparts. In other words, the SF-CNN models are manipulated by adversarial perturbations targeting SF-VGG11 \emph{less} than CNN models. Indeed, Grad-CAM visualizations (\cref{fig:grad_cam_sf}) for transfer attacks from SF-VGG11 show similar trend as \cref{fig:grad_cam}, where the attention of SF-ResNet18 is less affected by the adversarial patterns than ResNet18. Such a result suggests that SF-CNN models are overall more robust against transfer attacks than CNN models. In addition, the lack of transferability across SF-CNNs indicates that different features are learned by SF-CNN models using different architectures. As such, SF-CNNs present a unique advantage under the black-box setting: while an adversary may choose any CNN architecture as a surrogate model to generate a successful transfer attack, the same adversary is required to select the correct baseline architecture when facing an SF-CNN model.     

\section{Further analysis of SF Layer}
\label{sec:5_qual}
 \begin{figure}[t]
    \begin{subfigure}[t]{.24\linewidth}
        \centering
        \includegraphics[width=1\linewidth]{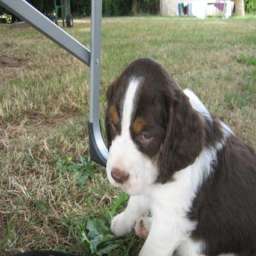}
        \includegraphics[width=1\linewidth]{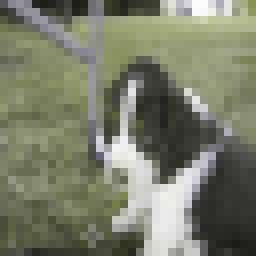}
        \includegraphics[width=1\linewidth]{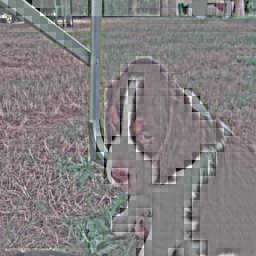}
    \end{subfigure}
    \begin{subfigure}[t]{.24\linewidth}
        \centering
        \includegraphics[width=1\linewidth]{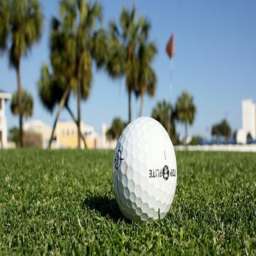}
        \includegraphics[width=1\linewidth]{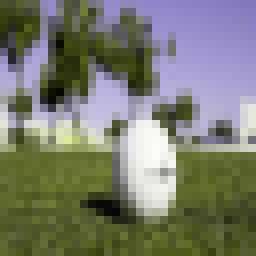}
        \includegraphics[width=1\linewidth]{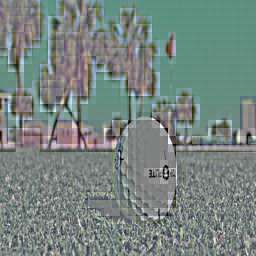}
    \end{subfigure}
    \begin{subfigure}[t]{.24\linewidth}
        \centering
        \includegraphics[width=1\linewidth]{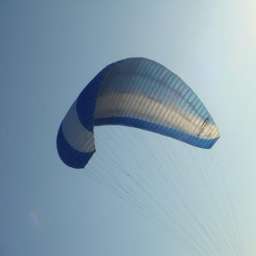}
        \includegraphics[width=1\linewidth]{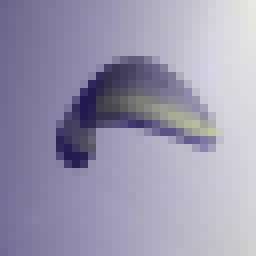}
        \includegraphics[width=1\linewidth]{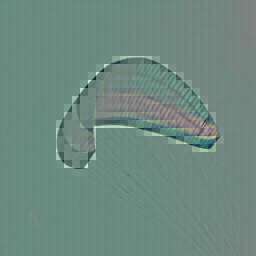}
    \end{subfigure}
    \begin{subfigure}[t]{.24\linewidth}
        \centering
        \includegraphics[width=1\linewidth]{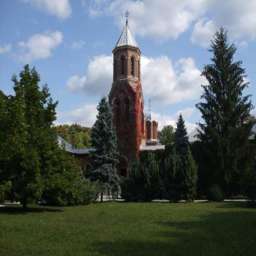}
        \includegraphics[width=1\linewidth]{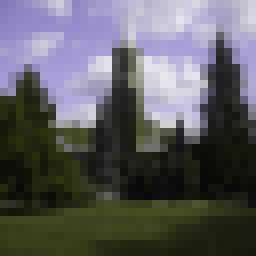}
        \includegraphics[width=1\linewidth]{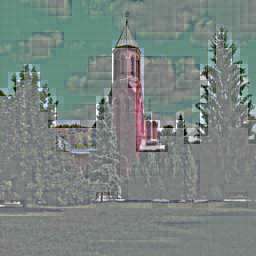}
    \end{subfigure}
    \caption{Examples of the original image (top), \textsc{lfr} (middle) and \textsc{hfr} (bottom). Despite using only the lowest frequency components, \textsc{lfr}s are visually more similar to the original images than \textsc{hfr}s.}
    \label{fig:frequency_recon}
\end{figure}

\begin{table}[t]\small
    \centering
    \resizebox{\linewidth}{!}{

    \begin{tabular}{lccc|ccc}
    \toprule
        model & \textsc{all} &\textsc{hfr}  & \textsc{lfr}  & \textsc{all-adv} & \textsc{hfr-adv}  & \textsc{lfr-adv} \\
         \midrule
        SF & $80.8\%$ &  $25.8\%$ & $54.6\%$ &  $47.2\%$ & $20.3\%$ & $39.9\%$\\
        
        $C_{88}$ & $79.6\%$  & $42.9\%$ & $49\%$ & $5.8\%$& $14.3\%$ & $45.2\%$\\
    \bottomrule
    \end{tabular}
    }
    \caption{Comparing SF-ResNet18 and $C_{88}\dash$ResNet18 model accuracy on high or low frequency reconstructions of clean or adversarial examples ($\epsilon=0.01$) of imagenette.}
    \label{tab:frequency_recon_result}
\end{table}

    In this section, we analyze the effect of the initial feature extraction layer by comparing the SF layer with an equivalently sized convolutional layer $C_{88}$. Concretely, $C_{88}$ layer is an $8\times8$ stride $8$ convolutional layer with $3$ input and $192$ output channels. 
    % \jy{Thus, the SF layer can be formulated as a $C_{88}$ layer with the specific $\mathbf{K}$ values in \cref{eqn:DCT} as kernel weights}.\footnote{The detail scheme is placed in \cref{apx:SPF_as_C88} due to space constraint.} 
    For a fair comparison, we construct a $C_{88}\dash$ResNet18 with identical architecture as SF-ResNet18 except for the first feature extraction layer.

    % we perform qualitative experiments to probe the apparent robustness of SF-CNN models. In particular, we 
    % with block-wise DCT process on $8\times8$ non-overlaping pixel patches and subsequent zigzag flattening 

    % As shown in~\cref{fig:frequency_recon}, we utilize an inverse function of the SPF layer to reconstruct an image with the lower (or higher) half of the frequency components are set to zero. Note that since we follow a block-wise DCT process, the reconstructed images also presents a block-wise pattern. High frequency reconstruction results in sharp edges within the constituent blocks while low frequency reconstructions results in smooth blocks that appears monochromatic.
    
\subsection{The impact of image frequency}
We compare SF-ResNet18's and $C_{88}\dash$ResNet18's responses to the frequency-based reconstruction of clean and adversarial images~\cite{wang2020high}. As shown in \cref{fig:frequency_recon}, we separate the lowest frequency, \ie, $\mathbf{D}_{0,0}$ in \cref{eqn:DCT}, for each pixel patch and color of the original image (\textsc{all}) to reconstruct the low-frequency image (\textsc{lfr}), and all other frequencies for the high-frequency image (\textsc{hfr}). We repeat the same process on the adversarial examples (\textsc{all-adv}) for the low- (\textsc{lfr-adv}) and high-frequency reconstructions (\textsc{hfr-adv}). 

In \Cref{tab:frequency_recon_result}, we find that $C_{88}\dash$ResNet18 shows higher accuracy on \textsc{hfr}, which indicates that the model relies more on the high-frequency information. In contrast, an SF-ResNet18 appears to utilize the lowest frequency component more, since the model has higher accuracy on \textsc{lfr}. Comparing \textsc{hfr-adv} and \textsc{lfr-adv}, both models obtain weaker performance with \textsc{hfr-adv} than \textsc{lfr-adv}, indicating that the harm of adversarial examples comes more from their high-frequency components than the low-frequency components. Since SF-ResNet18 utilizes more low-frequency patterns, it remains more robust than $C_{88}\dash$ResNet18 against adversarial attacks. Note that SF-ResNet18 shows better accuracy over $C_{88}$-ResNet18 in \textsc{all-adv} by utilizing the information in \textsc{lfr-adv}. In \cref{fig:frequency_recon}, \textsc{lfr} is visually more similar to the original image than \textsc{hfr} for a human observer.
    
\subsection{Mixture models of SF and $\mathbf{C_{88}}$ layers}
    To validate the robustness of different feature extractors, we conduct additional experiments with two types of mixture models between SF-ResNet18 and $C_{88}\dash$ResNet18.
    \subsubsection{Interpolation model}
       
% \begin{table}[t]
%     \centering
%     \resizebox{\linewidth}{!}{
%     \begin{tabular}{lccccc}
%     \toprule
%         $\alpha$ & $0$ & $0.25$  & $0.5$ & $0.75$  & $1$\\
%         \midrule
%         Acc. & $80.8\%$ & $79.0\%$ & $78.2\%$ & $78.8\%$ & $79.6\%$  \\
%         \hline
%         $\epsilon=0.003$ & $67.2\%$  & $64.8\%$   & $62.4\%$ & $60.0\%$  & $47.2\%$\\
%         $\epsilon=0.01$ & $31.2\%$ & $28.8\%$ & $24.0\%$& $21.2\%$ & $5.8\%$ \\
%         \bottomrule
%     \end{tabular}
%     }
%     \caption{The test accuracy of attacked image in interpolation models over different $\alpha$.}
%     \label{tab:ablation_DCTvsC88}
% \end{table}

% !! the above alpha is reversed!!

\begin{table}[t]\small
    \centering
    \resizebox{\linewidth}{!}{
    \begin{tabular}{lccccc}
    \toprule
        $\alpha$ & $0$ & $1/4$  & $1/2$ & $3/4$  & $1$\\
        \midrule
        Acc. & $79.6\%$ & $78.8\%$ & $78.2\%$ & $79.0\%$ &  $80.8\%$  \\
        \hline
        $\epsilon=0.003$ & $47.2\%$ & $60.0\%$   & $62.4\%$ & $64.8\%$  & $67.2\%$\\
        $\epsilon=0.01$ & $5.8\%$ & $21.2\%$ & $24.0\%$& $28.8\%$ & $31.2\%$ \\
        \bottomrule
    \end{tabular}
    }
    \caption{Performances of the interpolation models with different $\alpha$ values.}
    \label{tab:interpolation}
\end{table}

\begin{table}[t]\small
    \centering
    \resizebox{\linewidth}{!}{
    \begin{tabular}{lccccc}
    \toprule
        $\beta$ & $0$ & $1/64$ & $1/16$ & $1/4$ & $1$\\
        \midrule
        Acc. & $79.6\%$ & $79.4\%$ & $78.6\%$ & $78.6\%$ & $80.8\%$  \\
        \hline
        $\epsilon=0.003$ &  $47.2\%$ & $65.8\%$ & $65.6\%$ & $66.0\%$ & $67.2\%$  \\
        $\epsilon=0.01$ &  $5.8\%$ & $26.4\%$ & $26.8\%$ & $30.8\%$ & $31.2\%$  \\
        \bottomrule
    \end{tabular}
    }
    \caption{Performances of the substitution models with different $\beta$ values.}
    \label{tab:substitution}
\end{table}

% \begin{table}[t]
%     \centering
%     \resizebox{\linewidth}{!}{
%     \begin{tabular}{lcccccc}
%     \toprule
%         $\beta$ & $0$ & $0.01$ &$0.25$ & $0.5$ & $0.75$ & $1$\\
%         \midrule
%         Acc. & $79.6\%$ & $79.4\%$ & $\%$ & $78.6\%$ & $\%$ & $80.8\%$  \\
%         \hline
%         $\epsilon=0.003$ & $65.8\%$ & $47.2\%$ & $\%$ & $66.0\%$ & $\%$ & $67.2\%$  \\
%         $\epsilon=0.01$ & $5.8\%$ & $26.4\%$ & $\%$ & $30.8\%$ & $\%$ & $31.2\%$  \\
%         \bottomrule
%     \end{tabular}
%     }
%     \caption{The test accuracy of attacked image in substitution models of different $\beta$.}
%     \label{tab:substitution}
% \end{table}
   
% !! below beta meaning reversed!!
% \begin{table}[t]
%     \centering
%     \resizebox{\linewidth}{!}{
%     \begin{tabular}{lccccc}
%     \toprule
%         $\beta$ & $192$ & $48$ & $12$ & $3$ & $0$\\
%         \midrule
%         Acc. & $80.8\%$ & $78.6\%$ & $79.8\%$ & $79.4\%$ & $79.6\%$  \\
%         \hline
%         $\epsilon=0.003$ &  $67.2\%$ & $66.0\%$ & $65.6\%$ & $65.8\%$ & $47.2\%$  \\
%         $\epsilon=0.01$ &  $31.2\%$ & $30.8\%$ & $26.8\%$ & $26.4\%$ & $5.8\%$  \\
%         \bottomrule
%     \end{tabular}
%     }
%     \caption{The test accuracy of attacked image in substitution model models over different $\beta$.}
%     \label{tab:substitution}
% \end{table}
    
\begin{figure}[t]
    \begin{subfigure}[t]{.49\linewidth}
    \centering
    \includegraphics[width=\linewidth]{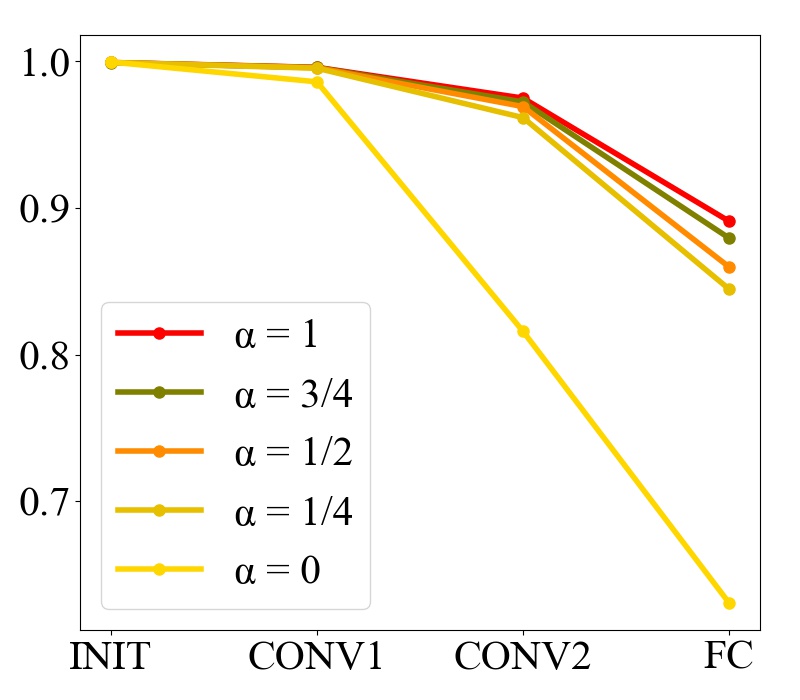}
    \caption{interpolation models}
    \label{subfig:cos_sim_alpha}
    \end{subfigure}
    \begin{subfigure}[t]{.49\linewidth}
    \centering
    \includegraphics[width=\linewidth]{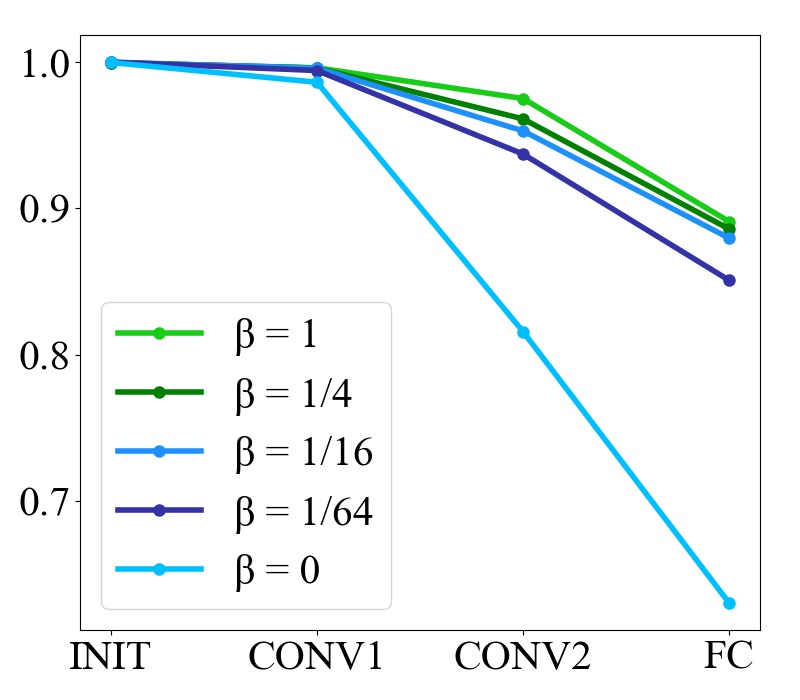}
    \caption{substitution models}
    \label{subfig:cos_sim_beta}
    \end{subfigure}
  \caption{The average cosine similarity of each layer outputs between clean images and the corresponding adversarial examples. Higher cosine similarity indicates less effect from the adversarial perturbation on the layer activation, corresponding to better adversarial robustness.}
  \label{fig:cos_sim_alpha}
\end{figure}

    First, we introduce an \emph{interpolation model} between SF-ResNet18 and $C_{88}\dash$ResNet18 by interpolating the first layer between the SF layer and the $C_{88}$ layer as follows:
    \begin{equation}\small
    \label{eqn:interpolation}
        \mathbf{K}_{inter} =  \alpha \mathbf{K}_{SF} + (1-\alpha) \mathbf{K}_{88} ,\,\,\alpha\in[0,1],
    \end{equation}
    where $\mathbf{K}_{88}$ and $\mathbf{K}_{SF}$ denote the kernel weights of $C_{88}$ layer and  $SF$ layer, respectively. As such, the interpolation model begins as $C_{88}$ model and gradually becomes an SF-model as the portion $\alpha$ increases.

    \Cref{tab:interpolation} shows that the interpolation models are most robust when $\alpha = 1$ (corresponding to SF-ResNet18), and least robust when $\alpha = 0$ (corresponding to $C_{88}\dash$ResNet18). The accuracy of the interpolation model consistently improves as $\alpha$ increases from $0$ to $1$. It manifests that the SF layer could extract more robust features than the $C_{88}$ layer. To further understand the effect of feature extraction, we measure the average cosine similarity between the layer outputs of clean and adversarial images at 1) the first layer \textsc{init} (\ie, \cref{eqn:interpolation}), 2) the first residual block \textsc{conv1}, 3) the second residual block \textsc{conv2}, and 4) the final fully connected layer \textsc{fc} (see \cref{apx:A1_architect} for detail). 
    
    The result in \cref{subfig:cos_sim_alpha} suggests that the adversarial attack process locates an adversarial pattern that is amplified along each consecutive layer. Since the adversarial pattern is bounded by an $\epsilon$ limit, the cosine similarity between the adversarial image and the clean image is approximately $1$. Then, after being processed by the initial feature extraction layer (\textsc{init}) and the first convolutional block (\textsc{conv1}), the layer output for the clean and adversarial examples begin to differ, leading to a decrease in their cosine similarity. Subsequent layer processes (\textsc{conv2} and \textsc{fc}) show even larger deviations.
    Note that while the cosine similarity at the last layer maintains at $0.89$ for SF-ResNet18 ($\alpha = 1$), the corresponding value dropped to $0.63$ for $C_{88}\dash$ResNet18 ($\alpha=0$). In particular, the cosine similarity in the last layer \textsc{fc} of the $\alpha=0$ model is significantly lower than other interpolation models, while $\alpha=0$ also corresponds to the interpolation model with no contributions from the SF layer. Thus, we conclude that training on a \emph{good} initial feature extraction layer, \ie, the SF layer leads to a more robust model weight in the subsequent layers, such that the layer output differences caused by the adversarial perturbations are less severe.

\subsubsection{Substitution model}

    To further evaluate the effect of each channel in SF layer, we propose the \emph{substitution model} which is constructed by replacing the kernels of the $C_{88}$ layer with that of the SF layer by the portion $\beta$, which can be defined as follows:
    \begin{equation}\small
    \label{eqn:substitution}
        \mathbf{K}_{sub} = concat(\mathbf{K}_{SF}[:\beta\cdot 192], \mathbf{K}_{88}[\beta\cdot 192:]), \beta \in [0,1],
    \end{equation}
    where $concat(\cdot)$ denotes the concatenation operation of two matrices. Thus, our $\mathbf{K}_{sub}$ has $192$ channels, which is identical to both the SF layer and the $C_{88}$ layer. Note that we sort the SF layer kernels from the low frequency to the high frequency, \ie, $\mathbf{K}_{SF}[0]$ is the lowest frequency channel, and $\mathbf{K}_{SF}[191]$ represents the highest frequency. As $\beta$ increases, we gradually replace the kernels in $C_{88}$ with the low-frequency kernels from the SF layer. 
    
    \Cref{tab:substitution} shows a similar trend as the interpolation models in \Cref{tab:interpolation}, where the substitution model is most robust when $\beta=1$ (corresponding to SF-ResNet18), and least robust when $\beta=0$ (corresponding to $C_{88}\dash$ResNet18). After adding the lowest frequency channel, \ie, with $\beta = 1/64$, the model performance significantly improves by $39.4\%$ and $355.6\%$ in terms of test accuracy for $\epsilon = 0.003$ and $\epsilon = 0.001$, respectively. Afterward, the model continues to improve as $\beta$ increases. As an additional experiment, adding the $1/4$ highest frequency channels of SF layer shows less promising results of $54.4\%$ with $\epsilon = 0.003$ and $10.1\%$ with $\epsilon = 0.01$. In contrast, adding $1/4$ lowest frequency channels of SF layer finds $66.0\%$ ($\epsilon = 0.003$) and $30.8\%$ ($\epsilon = 0.01$). Such comparison manifests that low-frequency components of the SF layer significantly contribute to the adversarial robustness of SF-CNN models. Moreover, we also measure the cosine similarities between model responses of clean and perturbed images for the substitution models in \cref{subfig:cos_sim_beta}, and observe the same cosine similarity drop along the layers. Moreover, the cosine similarity of the last layer \textsc{FC} of $\beta=0$ is significantly lower than other $\beta$ values. By comparing $\beta = 0$ with $\beta=1/64$ in both \cref{subfig:cos_sim_beta} and \Cref{tab:substitution}, we observe that substituting the \emph{lowest frequencies} alone causes the most significant improvements in terms of adversarial robustness. Substituting higher frequency kernels in addition only amounts to slight improvements. Such an observation supports the hypothesis that SF-CNNs gain robustness by utilizing the lower frequency components, as suggested in \Cref{tab:frequency_recon_result}.

% 5) qualitative analysis of attack process
%     1) DCT vs CNN
%         - rounding error
%         - pattern DCT <-> 8*8; 8*8 <-> 3*3; DCT <-> 3*3

% Compare DCT <-> 8*8; 8*8 <-> 3*3; DCT <-> 3*3 for:
        % adv vs normal example layer activation cos sim
        % how to evaluate frequency attention?
        % non-activate neuron (zero after relu) 統計 ->
        % weight 極端值 統計 -> 去掉極端值
        % weight orthogonal 統計 -> 增加 orthogonal (disentangle learning)
        % 
        % -> receptive field...?
%         - orthogonal projection matrix
%         - parameter influence function (extreme weight)
%         - high frequency
%         - DCT + CNN 
% todo:
%         - adversarial limits causing trouble?

% Theory:
    % disentangle, random projection

\section{Conclusions}
This work presents an empirical study exploring the adversarial robustness of CNNs operating in the spatial frequency domain. We propose the SF layer, which extracts spatial frequency components, and replace the first few layers of popular CNN models with the SF layer to obtain the SF-CNNs. SF-CNNs are significantly more robust than their CNN counterparts under white-box and black-box attacks. Further comparisons between the SF layer and a trainable CNN layer manifest the importance of the initial feature extraction layer, where the low-frequency filters play a crucial row in the adversarial robustness. We believe our discoveries provide insights into the nature of adversarial examples and may be utilized in future work to develop defensive methods against adversarial attacks.

%   Lastly, by individually substituting kernels of different frequency from the SF layer into the $C_{88}$ layer, we observe that while all frequency components are beneficial, the lowest frequency results in the most improvement on robustness. 

{\small
    \bibliographystyle{latex/ieee_fullname}
    \bibliography{ref}

\begin{thebibliography}{10}\itemsep=-1pt

\bibitem{ahmed1974discrete}
Nasir Ahmed, T. Natarajan, and Kamisetty~R Rao.
\newblock Discrete cosine transform.
\newblock {\em IEEE transactions on Computers}, 100(1):90--93, 1974.

\bibitem{athalye2018obfuscated}
Anish Athalye, Nicholas Carlini, and David Wagner.
\newblock Obfuscated gradients give a false sense of security: Circumventing
  defenses to adversarial examples.
\newblock In {\em International conference on machine learning}, pages
  274--283. PMLR, 2018.

\bibitem{carlini2017towards}
Nicholas Carlini and David Wagner.
\newblock Towards evaluating the robustness of neural networks.
\newblock In {\em 2017 ieee symposium on security and privacy (sp)}, pages
  39--57. IEEE, 2017.

\bibitem{chen2016compressing}
Wenlin Chen, James Wilson, Stephen Tyree, Kilian~Q Weinberger, and Yixin Chen.
\newblock Compressing convolutional neural networks in the frequency domain.
\newblock In {\em Proceedings of the 22nd ACM SIGKDD international conference
  on knowledge discovery and data mining}, pages 1475--1484, 2016.

\bibitem{croce2020robustbench}
Francesco Croce, Maksym Andriushchenko, Vikash Sehwag, Edoardo Debenedetti,
  Nicolas Flammarion, Mung Chiang, Prateek Mittal, and Matthias Hein.
\newblock Robustbench: a standardized adversarial robustness benchmark.
\newblock {\em arXiv preprint arXiv:2010.09670}, 2020.

\bibitem{dapello2020simulating}
Joel Dapello, Tiago Marques, Martin Schrimpf, Franziska Geiger, David Cox, and
  James~J DiCarlo.
\newblock Simulating a primary visual cortex at the front of cnns improves
  robustness to image perturbations.
\newblock {\em Advances in Neural Information Processing Systems},
  33:13073--13087, 2020.

\bibitem{demontis2019adversarial}
Ambra Demontis, Marco Melis, Maura Pintor, Matthew Jagielski, Battista Biggio,
  Alina Oprea, Cristina Nita-Rotaru, and Fabio Roli.
\newblock Why do adversarial attacks transfer? explaining transferability of
  evasion and poisoning attacks.
\newblock In {\em 28th USENIX security symposium (USENIX security 19)}, pages
  321--338, 2019.

\bibitem{devalois1990spatial}
Russell~L DeValois, Karen~K De~Valois, and Karen~K DeValois.
\newblock {\em Spatial vision}.
\newblock Number~14. Oxford University Press on Demand, 1990.

\bibitem{dong2021towards}
Minjing Dong, Yunhe Wang, Xinghao Chen, and Chang Xu.
\newblock Towards stable and robust addernets.
\newblock {\em Advances in Neural Information Processing Systems},
  34:13255--13265, 2021.

\bibitem{ehrlich2019deep}
Max Ehrlich and Larry~S Davis.
\newblock Deep residual learning in the jpeg transform domain.
\newblock In {\em Proceedings of the IEEE International Conference on Computer
  Vision}, pages 3484--3493, 2019.

\bibitem{geirhos2018imagenettrained}
Robert Geirhos, Patricia Rubisch, Claudio Michaelis, Matthias Bethge, Felix~A.
  Wichmann, and Wieland Brendel.
\newblock Imagenet-trained {CNN}s are biased towards texture; increasing shape
  bias improves accuracy and robustness.
\newblock In {\em International Conference on Learning Representations}, 2019.

\bibitem{glorot2010understanding}
Xavier Glorot and Yoshua Bengio.
\newblock Understanding the difficulty of training deep feedforward neural
  networks.
\newblock In {\em Proceedings of the thirteenth international conference on
  artificial intelligence and statistics}, pages 249--256. JMLR Workshop and
  Conference Proceedings, 2010.

\bibitem{goodfellow2018defense}
Ian Goodfellow.
\newblock Defense against the dark arts: An overview of adversarial example
  security research and future research directions.
\newblock {\em arXiv preprint arXiv:1806.04169}, 2018.

\bibitem{goodfellow2014explaining}
Ian~J Goodfellow, Jonathon Shlens, and Christian Szegedy.
\newblock Explaining and harnessing adversarial examples.
\newblock In {\em International Conference on Learning Representations}, 2015.

\bibitem{gueguen2018faster}
Lionel Gueguen, Alex Sergeev, Ben Kadlec, Rosanne Liu, and Jason Yosinski.
\newblock Faster neural networks straight from jpeg.
\newblock {\em Advances in Neural Information Processing Systems}, 31, 2018.

\bibitem{he2016deep}
Kaiming He, Xiangyu Zhang, Shaoqing Ren, and Jian Sun.
\newblock Deep residual learning for image recognition.
\newblock In {\em Proceedings of the IEEE conference on computer vision and
  pattern recognition}, pages 770--778, 2016.

\bibitem{heo2021rethinking}
Byeongho Heo, Sangdoo Yun, Dongyoon Han, Sanghyuk Chun, Junsuk Choe, and
  Seong~Joon Oh.
\newblock Rethinking spatial dimensions of vision transformers.
\newblock In {\em Proceedings of the IEEE/CVF International Conference on
  Computer Vision}, pages 11936--11945, 2021.

\bibitem{imagenette}
Jeremy Howard.
\newblock Imagenette.

\bibitem{huang2017densely}
Gao Huang, Zhuang Liu, Laurens Van Der~Maaten, and Kilian~Q Weinberger.
\newblock Densely connected convolutional networks.
\newblock In {\em Proceedings of the IEEE conference on computer vision and
  pattern recognition}, pages 4700--4708, 2017.

\bibitem{huang2021fsdr}
Jiaxing Huang, Dayan Guan, Aoran Xiao, and Shijian Lu.
\newblock Fsdr: Frequency space domain randomization for domain generalization.
\newblock In {\em Proceedings of the IEEE/CVF Conference on Computer Vision and
  Pattern Recognition}, pages 6891--6902, 2021.

\bibitem{huang2021rda}
Jiaxing Huang, Dayan Guan, Aoran Xiao, and Shijian Lu.
\newblock Rda: Robust domain adaptation via fourier adversarial attacking.
\newblock In {\em Proceedings of the IEEE/CVF International Conference on
  Computer Vision}, pages 8988--8999, 2021.

\bibitem{ilyas2019adversarialnotbugs}
Andrew Ilyas, Shibani Santurkar, Dimitris Tsipras, Logan Engstrom, Brandon
  Tran, and Aleksander Madry.
\newblock Adversarial examples are not bugs, they are features.
\newblock {\em Advances in neural information processing systems}, 32, 2019.

\bibitem{inkawhich2019feature}
Nathan Inkawhich, Wei Wen, Hai~Helen Li, and Yiran Chen.
\newblock Feature space perturbations yield more transferable adversarial
  examples.
\newblock In {\em Proceedings of the IEEE/CVF Conference on Computer Vision and
  Pattern Recognition}, pages 7066--7074, 2019.

\bibitem{kang2021stable}
Qiyu Kang, Yang Song, Qinxu Ding, and Wee~Peng Tay.
\newblock Stable neural ode with lyapunov-stable equilibrium points for
  defending against adversarial attacks.
\newblock {\em Advances in Neural Information Processing Systems},
  34:14925--14937, 2021.

\bibitem{kingma2014adam}
Diederik~P Kingma and Jimmy Ba.
\newblock Adam: A method for stochastic optimization.
\newblock {\em arXiv preprint arXiv:1412.6980}, 2014.

\bibitem{krizhevsky2009learning}
Alex Krizhevsky, Geoffrey Hinton, et~al.
\newblock Learning multiple layers of features from tiny images.
\newblock 2009.

\bibitem{lavin2016fast}
Andrew Lavin and Scott Gray.
\newblock Fast algorithms for convolutional neural networks.
\newblock In {\em Proceedings of the IEEE conference on computer vision and
  pattern recognition}, pages 4013--4021, 2016.

\bibitem{liu2016delving}
Yanpei Liu, Xinyun Chen, Chang Liu, and Dawn Song.
\newblock Delving into transferable adversarial examples and black-box attacks.
\newblock {\em arXiv preprint arXiv:1611.02770}, 2016.

\bibitem{liu2022convnet}
Zhuang Liu, Hanzi Mao, Chao-Yuan Wu, Christoph Feichtenhofer, Trevor Darrell,
  and Saining Xie.
\newblock A convnet for the 2020s.
\newblock In {\em Proceedings of the IEEE/CVF Conference on Computer Vision and
  Pattern Recognition}, pages 11976--11986, 2022.

\bibitem{madry2018towards}
Aleksander Madry, Aleksandar Makelov, Ludwig Schmidt, Dimitris Tsipras, and
  Adrian Vladu.
\newblock Towards deep learning models resistant to adversarial attacks.
\newblock In {\em International Conference on Learning Representations}, 2018.

\bibitem{majaj2002role}
Najib~J Majaj, Denis~G Pelli, Peri Kurshan, and Melanie Palomares.
\newblock The role of spatial frequency channels in letter identification.
\newblock {\em Vision research}, 42(9):1165--1184, 2002.

\bibitem{martinez2003complex}
Luis~M Martinez and Jose-Manuel Alonso.
\newblock Complex receptive fields in primary visual cortex.
\newblock {\em The neuroscientist}, 9(5):317--331, 2003.

\bibitem{nilsback2008automated}
Maria-Elena Nilsback and Andrew Zisserman.
\newblock Automated flower classification over a large number of classes.
\newblock In {\em 2008 Sixth Indian Conference on Computer Vision, Graphics \&
  Image Processing}, pages 722--729. IEEE, 2008.

\bibitem{papernot2016transferability}
Nicolas Papernot, Patrick McDaniel, and Ian Goodfellow.
\newblock Transferability in machine learning: from phenomena to black-box
  attacks using adversarial samples.
\newblock {\em arXiv preprint arXiv:1605.07277}, 2016.

\bibitem{qin2021fcanet}
Zequn Qin, Pengyi Zhang, Fei Wu, and Xi Li.
\newblock Fcanet: Frequency channel attention networks.
\newblock In {\em Proceedings of the IEEE/CVF international conference on
  computer vision}, pages 783--792, 2021.

\bibitem{quinlan1986induction}
J.~Ross Quinlan.
\newblock Induction of decision trees.
\newblock {\em Machine learning}, 1(1):81--106, 1986.

\bibitem{redmon2016you}
Joseph Redmon, Santosh Divvala, Ross Girshick, and Ali Farhadi.
\newblock You only look once: Unified, real-time object detection.
\newblock In {\em Proceedings of the IEEE conference on computer vision and
  pattern recognition}, pages 779--788, 2016.

\bibitem{sachs1971spatial}
Murray~B Sachs, Jacob Nachmias, and John~G Robson.
\newblock Spatial-frequency channels in human vision.
\newblock {\em JOSA}, 61(9):1176--1186, 1971.

\bibitem{selvaraju2017grad}
Ramprasaath~R Selvaraju, Michael Cogswell, Abhishek Das, Ramakrishna Vedantam,
  Devi Parikh, and Dhruv Batra.
\newblock Grad-cam: Visual explanations from deep networks via gradient-based
  localization.
\newblock In {\em Proceedings of the IEEE international conference on computer
  vision}, pages 618--626, 2017.

\bibitem{shafahi2019adversarial}
Ali Shafahi, Mahyar Najibi, Mohammad~Amin Ghiasi, Zheng Xu, John Dickerson,
  Christoph Studer, Larry~S Davis, Gavin Taylor, and Tom Goldstein.
\newblock Adversarial training for free!
\newblock {\em Advances in Neural Information Processing Systems}, 32, 2019.

\bibitem{shen2021dct}
Xing Shen, Jirui Yang, Chunbo Wei, Bing Deng, Jianqiang Huang, Xian-Sheng Hua,
  Xiaoliang Cheng, and Kewei Liang.
\newblock Dct-mask: Discrete cosine transform mask representation for instance
  segmentation.
\newblock In {\em Proceedings of the IEEE/CVF Conference on Computer Vision and
  Pattern Recognition}, pages 8720--8729, 2021.

\bibitem{simonyan2014very}
Karen Simonyan and Andrew Zisserman.
\newblock Very deep convolutional networks for large-scale image recognition.
\newblock {\em arXiv preprint arXiv:1409.1556}, 2014.

\bibitem{stutz2019disentangling}
David Stutz, Matthias Hein, and Bernt Schiele.
\newblock Disentangling adversarial robustness and generalization.
\newblock In {\em Proceedings of the IEEE/CVF Conference on Computer Vision and
  Pattern Recognition}, pages 6976--6987, 2019.

\bibitem{szegedy2013intriguing}
Christian Szegedy, Wojciech Zaremba, Ilya Sutskever, Joan Bruna, Dumitru Erhan,
  Ian Goodfellow, and Rob Fergus.
\newblock Intriguing properties of neural networks.
\newblock In {\em International Conference on Learning Representations}, 2014.

\bibitem{tan2021efficientnetv2}
Mingxing Tan and Quoc Le.
\newblock Efficientnetv2: Smaller models and faster training.
\newblock In {\em International Conference on Machine Learning}, pages
  10096--10106. PMLR, 2021.

\bibitem{tramer2020adaptive}
Florian Tramer, Nicholas Carlini, Wieland Brendel, and Aleksander Madry.
\newblock On adaptive attacks to adversarial example defenses.
\newblock {\em Advances in Neural Information Processing Systems},
  33:1633--1645, 2020.

\bibitem{tsipras2018robustness}
Dimitris Tsipras, Shibani Santurkar, Logan Engstrom, Alexander Turner, and
  Aleksander Madry.
\newblock Robustness may be at odds with accuracy.
\newblock In {\em International Conference on Learning Representations}, 2019.

\bibitem{tu2019autozoom}
Chun-Chen Tu, Paishun Ting, Pin-Yu Chen, Sijia Liu, Huan Zhang, Jinfeng Yi,
  Cho-Jui Hsieh, and Shin-Ming Cheng.
\newblock Autozoom: Autoencoder-based zeroth order optimization method for
  attacking black-box neural networks.
\newblock In {\em Proceedings of the AAAI Conference on Artificial
  Intelligence}, volume~33, pages 742--749, 2019.

\bibitem{vuilleumier2003distinct}
Patrik Vuilleumier, Jorge~L Armony, Jon Driver, and Raymond~J Dolan.
\newblock Distinct spatial frequency sensitivities for processing faces and
  emotional expressions.
\newblock {\em Nature neuroscience}, 6(6):624--631, 2003.

\bibitem{vuyyuru2020biologically}
Manish~Reddy Vuyyuru, Andrzej Banburski, Nishka Pant, and Tomaso Poggio.
\newblock Biologically inspired mechanisms for adversarial robustness.
\newblock {\em Advances in Neural Information Processing Systems},
  33:2135--2146, 2020.

\bibitem{wallace1992jpeg}
Gregory~K Wallace.
\newblock The jpeg still picture compression standard.
\newblock {\em IEEE transactions on consumer electronics}, 38(1):xviii--xxxiv,
  1992.

\bibitem{wang2020high}
Haohan Wang, Xindi Wu, Zeyi Huang, and Eric~P Xing.
\newblock High-frequency component helps explain the generalization of
  convolutional neural networks.
\newblock In {\em Proceedings of the IEEE/CVF Conference on Computer Vision and
  Pattern Recognition}, pages 8684--8694, 2020.

\bibitem{wong2020fast}
Eric Wong, Leslie Rice, and J~Zico Kolter.
\newblock Fast is better than free: Revisiting adversarial training.
\newblock {\em arXiv preprint arXiv:2001.03994}, 2020.

\bibitem{wu2020adversarial}
Dongxian Wu, Shu-Tao Xia, and Yisen Wang.
\newblock Adversarial weight perturbation helps robust generalization.
\newblock {\em Advances in Neural Information Processing Systems},
  33:2958--2969, 2020.

\bibitem{xu2020learning}
Kai Xu, Minghai Qin, Fei Sun, Yuhao Wang, Yen-Kuang Chen, and Fengbo Ren.
\newblock Learning in the frequency domain.
\newblock In {\em Proceedings of the IEEE/CVF Conference on Computer Vision and
  Pattern Recognition}, pages 1740--1749, 2020.

\bibitem{zhang2019theoretically}
Hongyang Zhang, Yaodong Yu, Jiantao Jiao, Eric Xing, Laurent El~Ghaoui, and
  Michael Jordan.
\newblock Theoretically principled trade-off between robustness and accuracy.
\newblock In {\em International conference on machine learning}, pages
  7472--7482. PMLR, 2019.

\end{thebibliography}
}

\clearpage
\appendix
\renewcommand{\theequation}{\thesection.\arabic{equation}}
\setcounter{equation}{0}
\section{SF-CNN model details}
\label{apx:A1_architect}

\begin{figure}[ht]
        \centering
        \begin{subfigure}[t]{.4\linewidth}
            \centering
            \includegraphics[width=\linewidth]{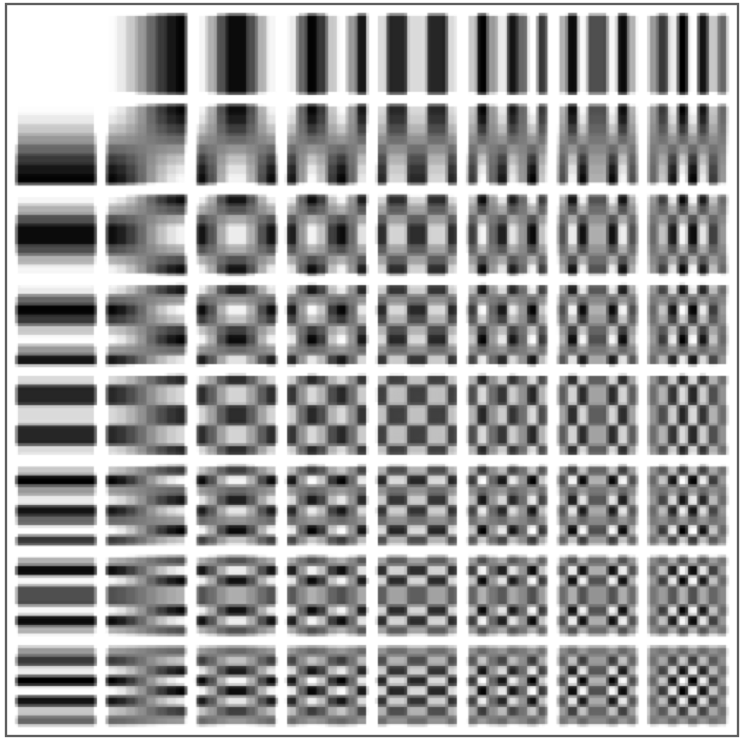}
        \end{subfigure}
        \begin{subfigure}[t]{.4\linewidth}
            \centering
            \includegraphics[width=\linewidth]{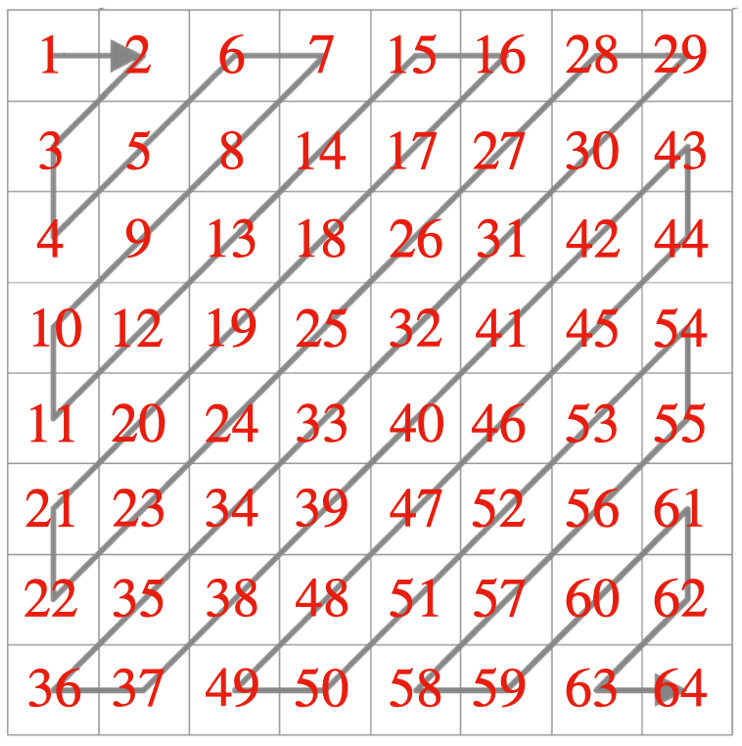}
        \end{subfigure}
        \caption{The DCT basis and the zigzag pattern}
        \label{apxfig:DCT_zigzag}
    \end{figure}

We present the detail layer-by-layer design of the experimented CNN and SF-CNN model architectures in \Cref{tab:model_architecture}. Besides, the way to formulate the SF layer as a $C_{88}$ layer are described as follows. A $C_{88}$ layer is a convolutional layer with $192$ instances of $3\times8\times8$ kernel filters. Since each kernel filter ``sees'' all three input channels (color space) at once, to create a convolutional layer with the same function as \cref{eqn:DCT_kernel}, which only processes one color, we fill two $8\times8$ layers with $0$ and one layer with the $\mathbf{K}^{u,v}$ matrix with a particular set of $u$ and $v$ value. Thus, $3$ color selections and $64$ possible ${u,v}$s results in $192$ kernel filters, matching the dimensions of the $C_{88}$ layer. Three kernels representing the same spatial frequency are grouped together and each frequency group is ordered by the zigzag scanning pattern shown in \cref{apxfig:DCT_zigzag}.

\setcounter{equation}{0}
\section{Additional Experiments}
%Follow this paper: https://openreview.net/pdf?id=8uWOTxbwo-Z
% Detecting AutoAttack Perturbations in the Frequency Domain
% or https://arxiv.org/pdf/2103.03000.pdf SpectralDefense: Detecting Adversarial Attacks on CNNs in the Fourier Domain

\subsection{Detection of adversarial example} 
\label{apx:detection}
\begin{figure}[t]
    \centering
    % RGB_freq
    \begin{subfigure}[t]{\linewidth}
        \centering
        \includegraphics[width=.45\linewidth]{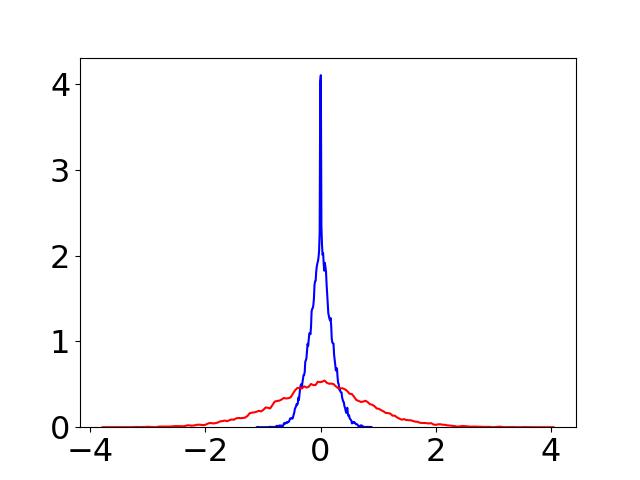}
        \includegraphics[width=.45\linewidth]{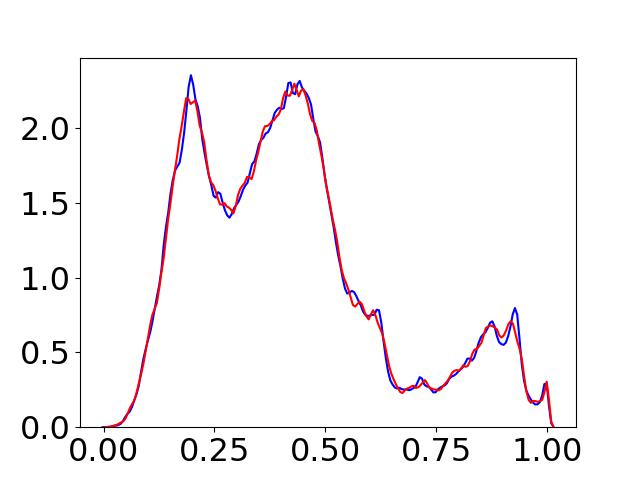}
        \caption{Adversarial attack in the pixel domain}
    \end{subfigure}
    
    \begin{subfigure}[t]{\linewidth}
        \centering
        \includegraphics[width=.45\linewidth]{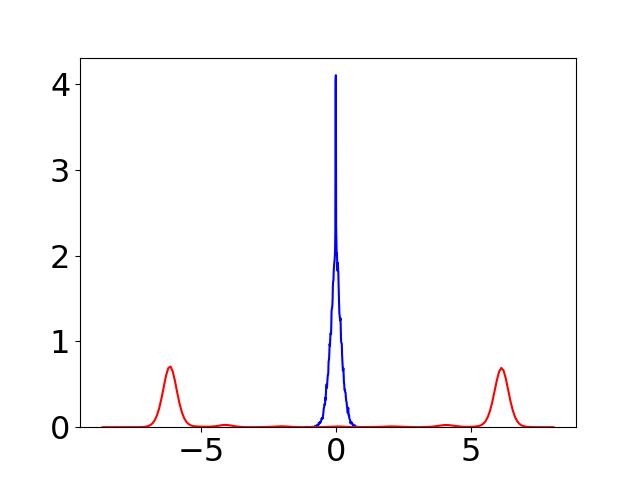}
        \includegraphics[width=.45\linewidth]{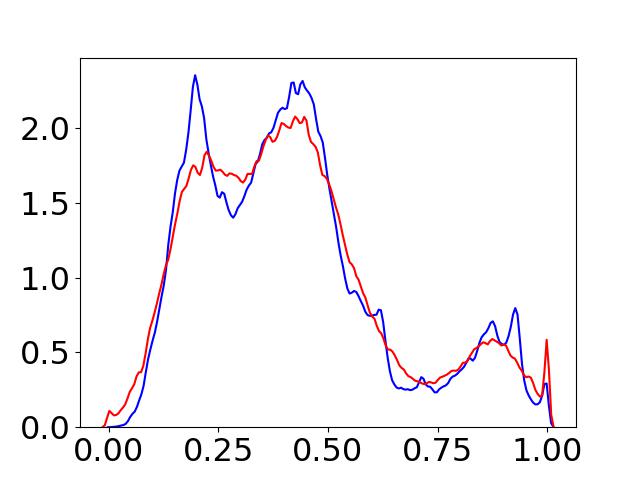}
        \caption{Adversarial attack in the frequency domain}
    \end{subfigure}
    \caption{The distributions of frequency component values (left) and pixel values (right) of an adversarial example (red) and the original test image (blue). For adversarial attack in both the pixel domain and the frequency domain, the frequency value distributions show clear differences while the pixel values are ambiguous.}
    \label{fig:adv_distribution}
\end{figure}

While the adversarial example can cause CNNs to misclassify and raise several concerns about models applicability, it is of paramount importance to develop detection methods for adversarial examples, thus allowing them to reject them. \cref{fig:adv_distribution} compares the distribution of frequency component values and that of the pixel values.\footnote{The distribution is plotted for an arbitrary image since all images show the same pattern between clean and adversarial examples.} We observe that frequency value distribution for natural images forms a peak centered at $0$. Noticeably, adversarial attacks in the pixel domain flatten the peak, whereas attacks in the frequency domain split into two peaks. In contrast, the same adversarial perturbations do not show simple patterns of change in the pixel values. Based on such observation, we preprocess the DCT coefficient of each image by dividing the range $[-3, 3]$ into $5$ intervals and count the frequency components with values in each interval as the input feature. Then, we train a Decision Tree classifier~\cite{quinlan1986induction} for detection, which is able to achieve $100\%$ detection accuracy across all models (including standard CNNs and SF-CNNs) and datasets since the frequency distribution shifts shown in \cref{fig:adv_distribution} are significant and universal.
\subsection{Evaluation with adversarial training}
\label{subsec:4-3advtrain}

% We compare SF-CNN and CNN models under adversarial training~\cite{madry2018towards}. Here w
We follow \cite{shafahi2019adversarial} to perform adversarial training with $\epsilon=0.03$ adversarial examples. The same adversarial training procedure is applied to both SF-ResNet18 and ResNet18. As shown in \Cref{tab:adv_train}, we observe that SF-CNNs are better or equal to CNNs under adversarial training while maintaining the same test accuracy. The result validates the efficacy of utilizing the frequency domain with adversarial training methods.

\begin{table}[ht]\small
    \centering
    \begin{tabular}{llccc}
    \toprule
    model & Acc. & $\epsilon=0.003$ & $\epsilon=0.01$ & $\epsilon=0.03$\\
    \midrule
    SF-ResNet18 & $76.2\%$ & $72.2\%$ & $65.4\%$ & $38.0\%$ \\
    ResNet18  &$76.2\%$ & $72.2\%$ & $63.4\%$ & $32.0\%$ \\
    \bottomrule 
    \end{tabular}
    \caption{Attacked accuracy of adversarially trained SF-ResNet18 and ResNet18 on the imagenette dataset. ``Acc.'' denotes clean test accuracy.}
    \label{tab:adv_train}
\end{table}

\begin{table*}[ht]
    \centering
\begin{tabular}{cc}
        ResNet18 & SF-ResNet18 \\
        \midrule
        $7\times7, 64, \text{stride} 2$ & \multirow{4}*{$\left.\rule{0cm}{1.55cm}\right\}$ \shortstack{replaced with \\the SF layer}}\\
        \cmidrule(lr){1-1}
        $3\times3$ max pool, stride $2$ & \\
        \cmidrule(lr){1-1}
        $\begin{bmatrix}3\times3, 64\\ 3\times3, 64\end{bmatrix}\times 2$ &\\
        \cmidrule(lr){1-1}
        $\begin{bmatrix}3\times3, 128\\ 3\times3, 128\end{bmatrix}\times 2$ &\\
        \cmidrule(lr){1-1}\cmidrule(lr){2-2}
        $\begin{bmatrix}3\times3, 256\\ 3\times3, 256\end{bmatrix}\times 2$ & $\begin{bmatrix}3\times3, 256\\ 3\times3, 256\end{bmatrix}\times 2$ \\
        \cmidrule(lr){1-1}\cmidrule(lr){2-2}
        $\begin{bmatrix}3\times3, 512\\ 3\times3, 512\end{bmatrix}\times 2$ & $\begin{bmatrix}3\times3, 512\\ 3\times3, 512\end{bmatrix}\times 2$\\
        \cmidrule(lr){1-1}\cmidrule(lr){2-2}
        $7\times7$ average pool & $7\times7$ average pool\\
        \cmidrule(lr){1-1}\cmidrule(lr){2-2}
        $512\times 10$ & $512\times 10$\\
    \end{tabular}
    \begin{tabular}{cc}
        EfficientNet & SF-EfficientNet \\
        \midrule
        $3\times3, 24, \text{stride} 2$ & \multirow{8}*{$\left.\rule{0cm}{1.55cm}\right\}$ \shortstack{replaced with \\the SF layer}}\\
        \cmidrule(lr){1-1}
        Fused-MBConv $3\times3$, 24 $\times3$ &\\
        \cmidrule(lr){1-1}
        Fused-MBConv $3\times3$, 48 $\times5$ &\\
        \cmidrule(lr){1-1}
        Fused-MBConv $3\times3$, 80 $\times5$ &\\
        \cmidrule(lr){1-1}
        MBConv $3\times3$, 160 $\times7$ &\\
        \cmidrule(lr){1-1}
        MBConv $3\times3$, 176 $\times14$ &\\
        \cmidrule(lr){1-1}\cmidrule(lr){2-2}
        MBConv $3\times3$, 304 $\times18$ & MBConv $3\times3$, 304 $\times18$\\
        \cmidrule(lr){1-1}\cmidrule(lr){2-2}
        MBConv $3\times3$, 512 $\times5$ & MBConv $3\times3$, 512 $\times5$\\
        \cmidrule(lr){1-1}\cmidrule(lr){2-2}
        $1\times1$,1792, stride 1 &  $1\times1$,1792, stride 1\\
        \cmidrule(lr){1-1}\cmidrule(lr){2-2}
        $1\times1$ average pool & $1\times1$ average pool\\
        \cmidrule(lr){1-1}\cmidrule(lr){2-2}
        $1792\times 10$ & $1792\times 10$\\
    \end{tabular}
    
    \vspace{1em}
    \begin{tabular}{cc}
        DenseNet & SF-DenseNet \\
        \midrule
        & SF layer \\
        \cmidrule(lr){1-1}\cmidrule(lr){2-2}
        $7\times7, 64, \text{stride} 2$ & $7\times7, 192, \text{stride} 2$ \\
        \cmidrule(lr){1-1}\cmidrule(lr){2-2}
        $3\times3$ max pool, stride $2$ & $3\times3$ max pool, stride $2$\\
        \cmidrule(lr){1-1}\cmidrule(lr){2-2}
        $\begin{bmatrix}1\times1, 128\\ 3\times3, 32\end{bmatrix}\times 6$ & $\begin{bmatrix}1\times1, 128\\ 3\times3, 32\end{bmatrix}\times 6$\\
        \cmidrule(lr){1-1}\cmidrule(lr){2-2}
        $1\times1, 128$ & $1\times1, 192$ \\
        \cmidrule(lr){1-1}\cmidrule(lr){2-2}
        $2\times2$ average pool, stride $2$ & $2\times2$ average pool, stride $2$\\
        \cmidrule(lr){1-1}\cmidrule(lr){2-2}
        $\begin{bmatrix}1\times1, 128\\ 3\times3, 32\end{bmatrix}\times 12$ & $\begin{bmatrix}1\times1, 128\\ 3\times3, 32\end{bmatrix}\times 12$\\
        \cmidrule(lr){1-1}\cmidrule(lr){2-2}
        $1\times1, 256$ & $1\times1, 288$ \\
        \cmidrule(lr){1-1}\cmidrule(lr){2-2}
        $2\times2$ average pool, stride $2$ & $2\times2$ average pool, stride $2$\\
        \cmidrule(lr){1-1}\cmidrule(lr){2-2}
        $\begin{bmatrix}1\times1, 128\\ 3\times3, 32\end{bmatrix}\times 24$ & $\begin{bmatrix}1\times1, 128\\ 3\times3, 32\end{bmatrix}\times 24$\\
        \cmidrule(lr){1-1}\cmidrule(lr){2-2}
        $1\times1, 512$ & $1\times1, 528$ \\
        \cmidrule(lr){1-1}\cmidrule(lr){2-2}
        $2\times2$ average pool, stride $2$ & $2\times2$ average pool, stride $2$\\
        \cmidrule(lr){1-1}\cmidrule(lr){2-2}
        $\begin{bmatrix}1\times1, 128\\ 3\times3, 32\end{bmatrix}\times 16$ & $\begin{bmatrix}1\times1, 128\\ 3\times3, 32\end{bmatrix}\times 16$\\
        \cmidrule(lr){1-1}\cmidrule(lr){2-2}
        $7\times7$ global average pool & $7\times7$ global average pool\\
        \cmidrule(lr){1-1}\cmidrule(lr){2-2}
        $1024\times 10$ & $1040\times 10$\\
    \end{tabular}
    \begin{tabular}{cc}
        VGG11 & SF-VGG11 \\
        \midrule
        $3\times3, 64, \text{stride} 2$ & \multirow{1}*{$\left.\rule{0cm}{1.2cm}\right\}$ \shortstack{replaced with \\the SF layer}}\\
        \cmidrule(lr){1-1}
        $2\times2$ max pool, stride $2$ & \\
        \cmidrule(lr){1-1}
        $3\times3$, 128, stride = 1 & \\
        \cmidrule(lr){1-1}
        $2\times2$ max pool, stride $2$ & \\
        \cmidrule(lr){1-1}\cmidrule(lr){2-2}
        $\begin{bmatrix}3\times3, 256\\ 3\times3, 256\end{bmatrix}\times 2$ & $\begin{bmatrix}3\times3, 256\\ 3\times3, 256\end{bmatrix}\times 2$\\
        \cmidrule(lr){1-1}\cmidrule(lr){2-2}
        $2\times2$ max pool, stride $2$ & $2\times2$ max pool, stride $2$\\
        \cmidrule(lr){1-1}\cmidrule(lr){2-2}
        $\begin{bmatrix}3\times3, 512\\ 3\times3, 512\end{bmatrix}\times 2$ & $\begin{bmatrix}3\times3, 512\\ 3\times3, 512\end{bmatrix}\times 2$ \\
        \cmidrule(lr){1-1}\cmidrule(lr){2-2}
        $2\times2$ max pool, stride $2$ & $2\times2$ max pool, stride $2$\\
        \cmidrule(lr){1-1}\cmidrule(lr){2-2}
        $\begin{bmatrix}3\times3, 512\\ 3\times3, 512\end{bmatrix}\times 2$ & $\begin{bmatrix}3\times3, 512\\ 3\times3, 512\end{bmatrix}\times 2$\\
        \cmidrule(lr){1-1}\cmidrule(lr){2-2}
        $2\times2$ max pool, stride $2$ & $2\times2$ max pool, stride $2$\\
        \cmidrule(lr){1-1}\cmidrule(lr){2-2}
        $4096\times 10$ & $4096\times 10$\\
    \end{tabular}
    
    \caption{The CNN and SF-CNN model architectures. Here, $k\times k, h$ indicate convolution layer (if not specified as a pooling layer) with $h$ output channels of $k\times k$ kernels with stride $1$ (if stride not specified). Brackets denote residual blocks. The last layers are fully connected layers.}
    \label{tab:model_architecture}
\end{table*}
\subsection{More Grad-cam visualization results}    
\label{apx:grad_cam}

\begin{figure*}[htp]
    \centering
    \begin{places}{\textwidth}{8}
        \place{Image}{
            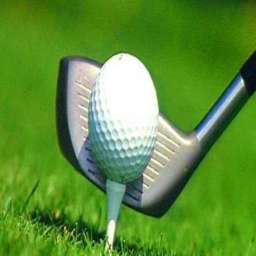,
            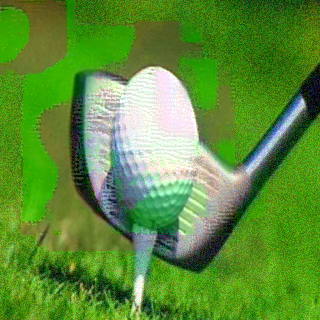,
            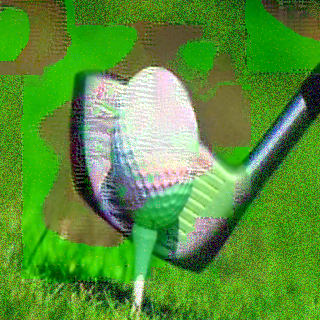,
            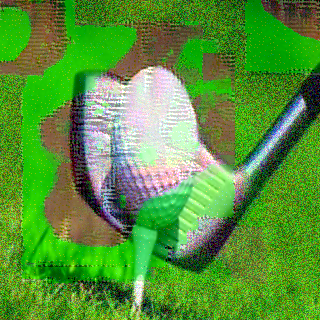,
            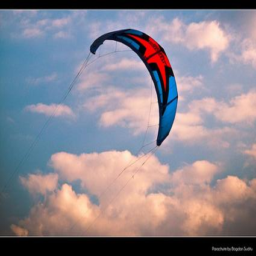,
            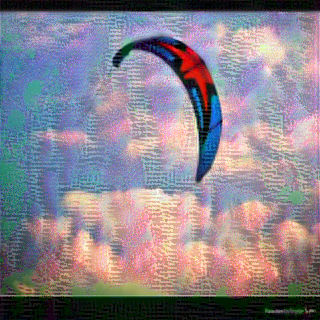,
            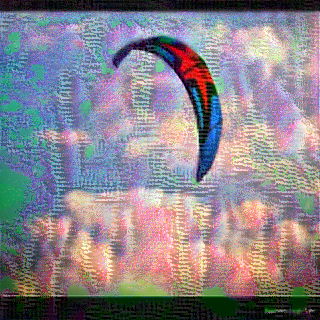,
            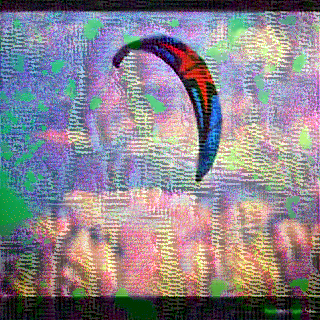
        }
        \place{}{}
        \place{ResNet18}{
            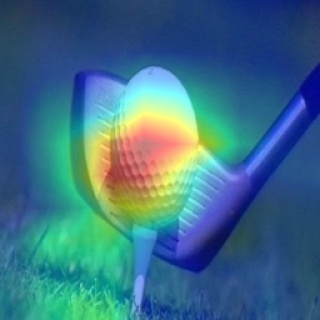,
            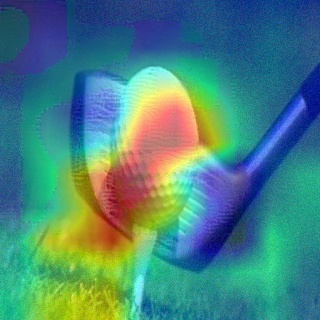,
            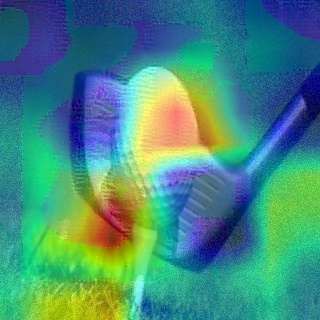,
            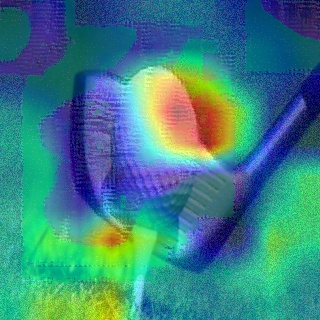,
            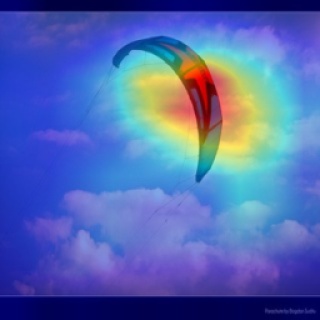,
            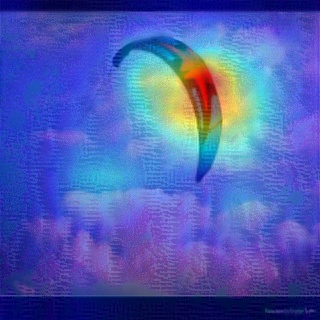,
            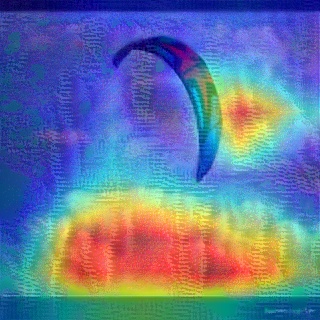,
            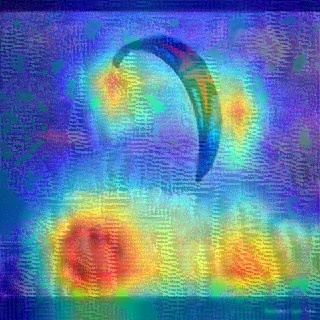
        }
        \place{EfficientNet}{
            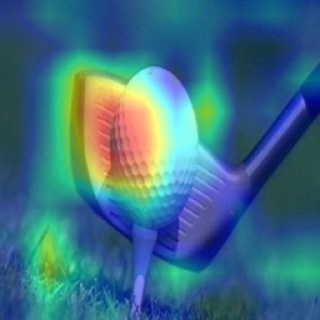,
            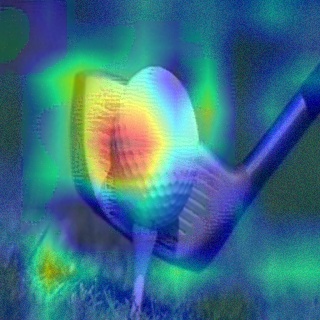,
            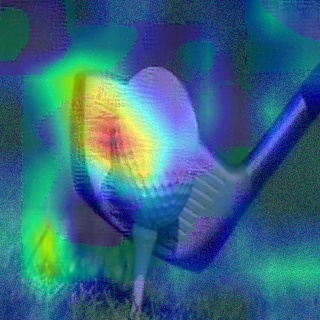,
            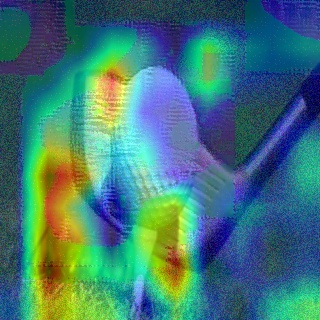,
            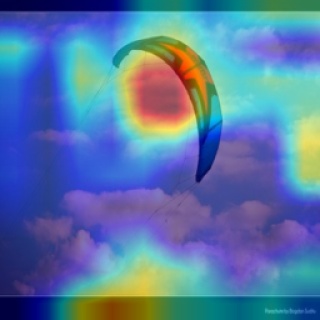,
            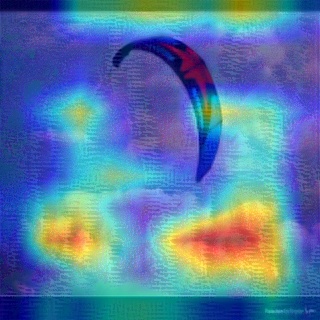,
            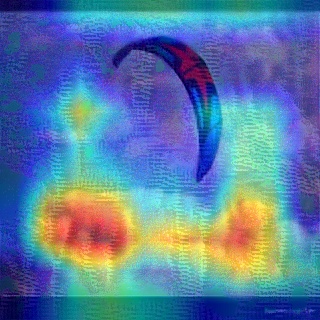,
            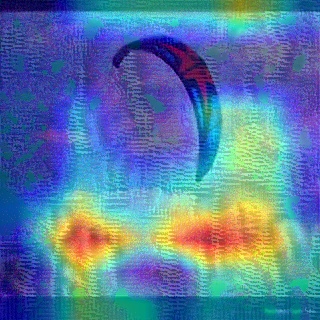,
        }
        \place{DenseNet}{
            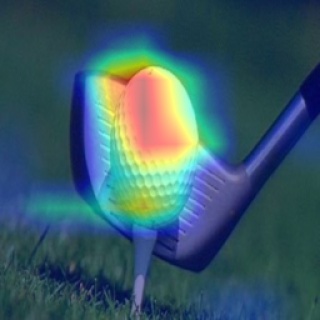,
            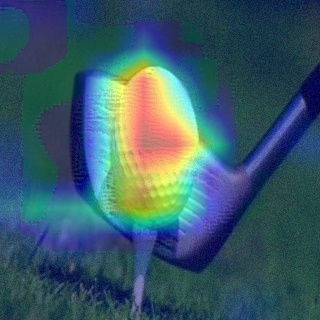,
            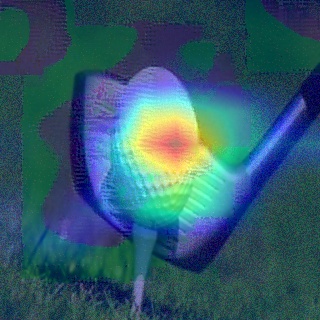,
            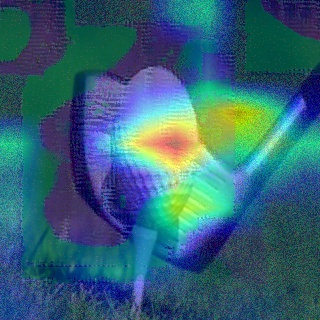,
            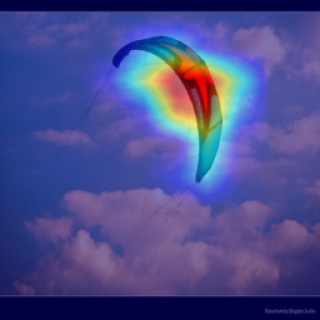,
            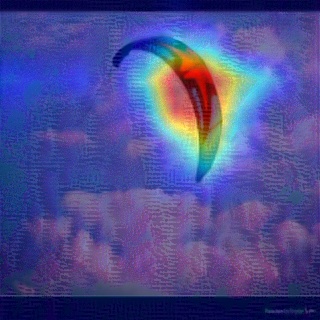,
            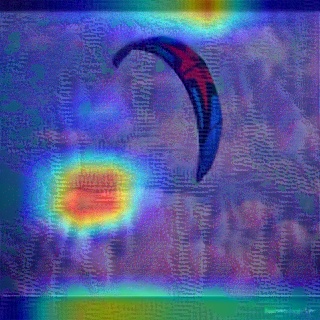,
            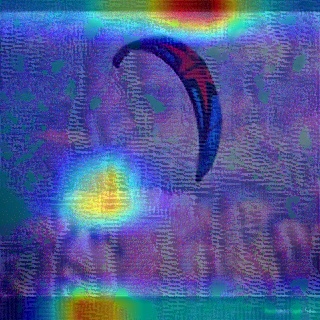,
        }
        \place{}{}
        \place{SF-ResNet18}{
            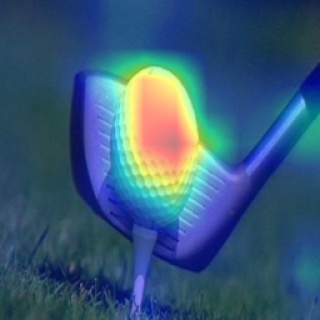,
            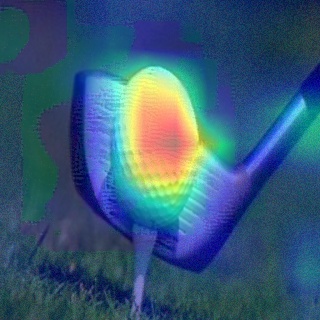,
            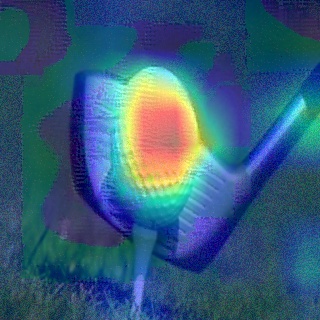,
            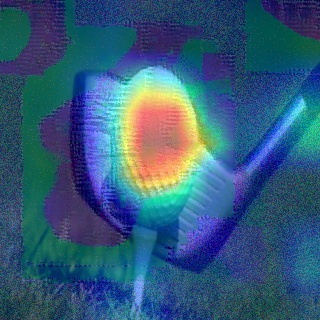,
            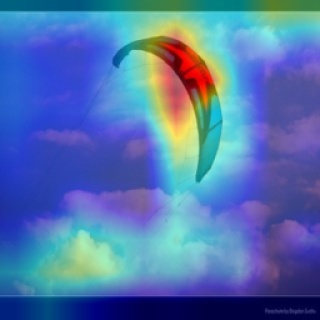,
            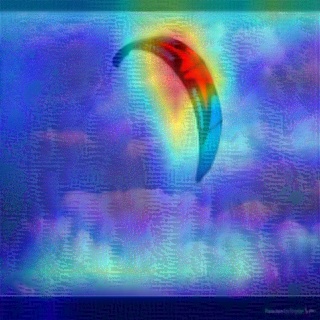,
            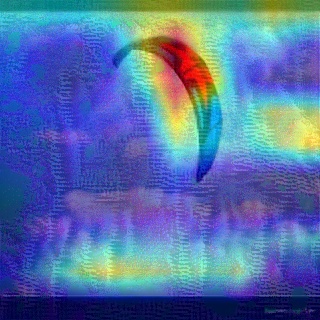,
            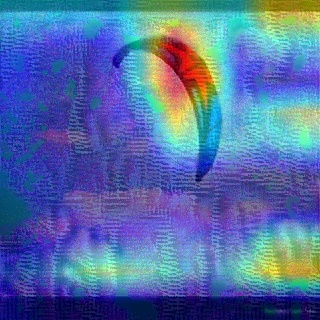
        }
        \place{SF-EfficientNet}{
            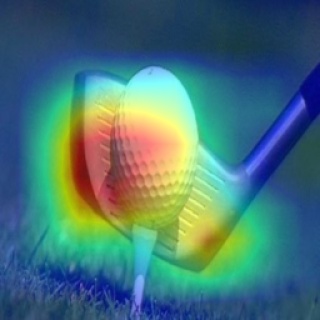,
            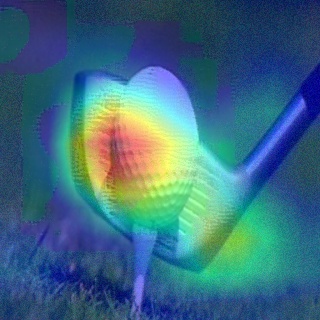,
            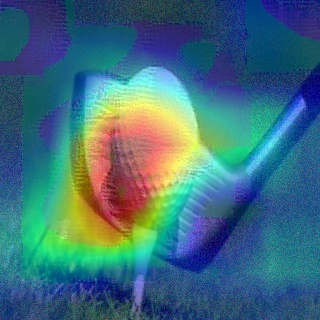,
            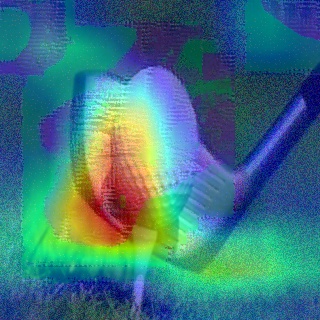,
            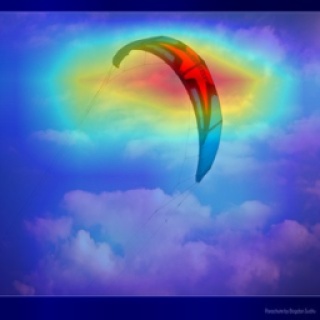,
            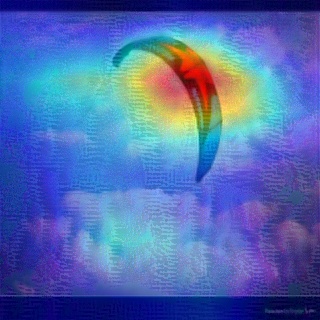,
            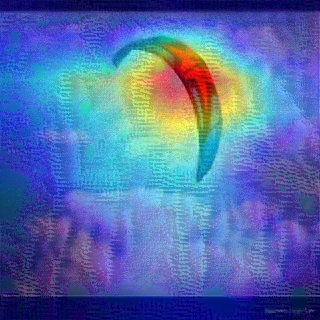,
            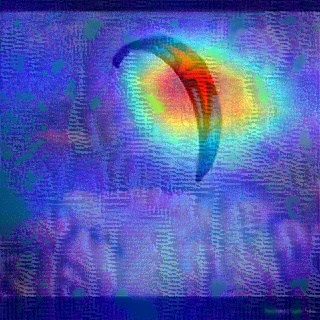
        }
        \place{SF-DenseNet}{
            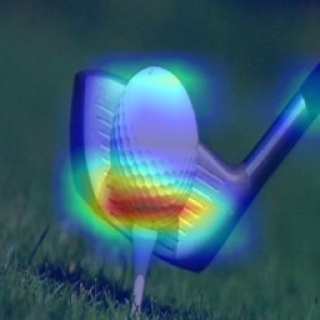,
            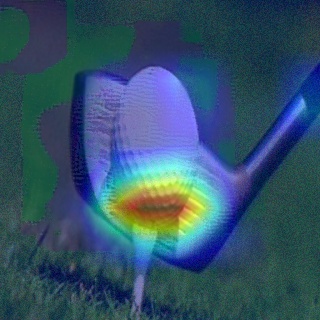,
            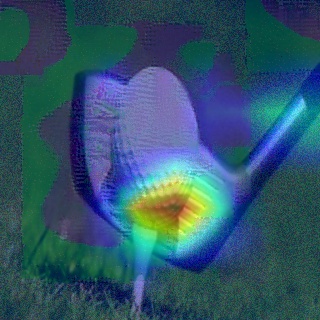,
            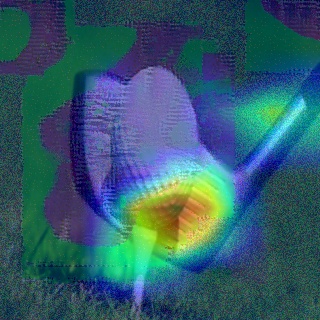,
            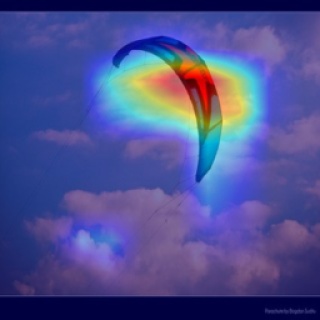,
            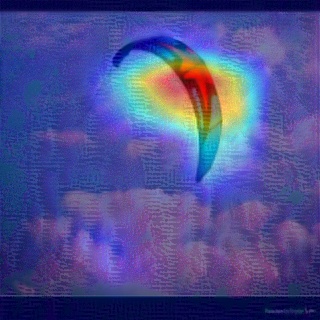,
            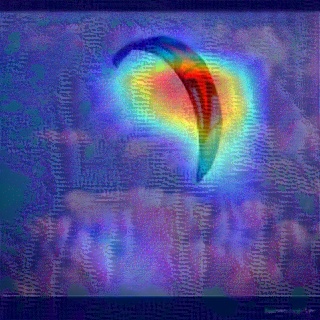,
            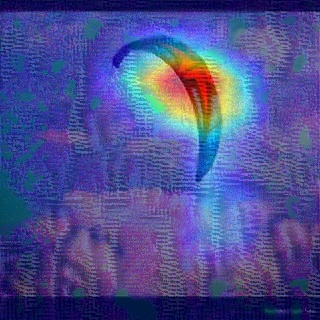,
        }
    \end{places}
    \caption{Grad-Cam visualizations of model attention on two adversarial examples generated from \textbf{VGG11}. Each row shows the progression from $\epsilon=0$ (not attacked) to $\epsilon=0.1$, $0.2$ and $0.3$ from left to right.}
    \label{apxfig:grad_cam}
\end{figure*}

\begin{figure*}[htp]
    \centering
    \begin{places}{\textwidth}{8}
        \place{Image}{
            figure/apx/golf.jpg,
            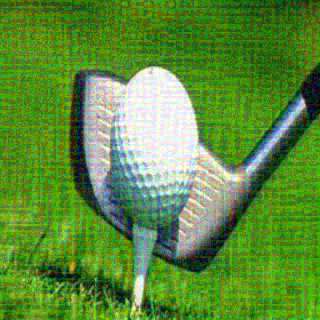,
            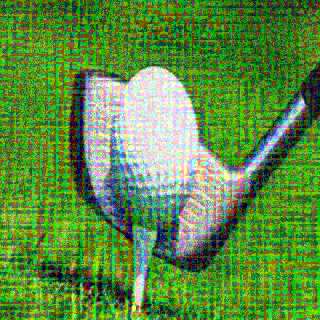,
            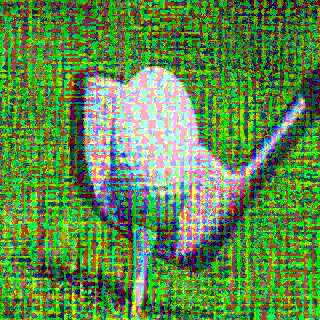,
            figure/apx/chute.jpg,
            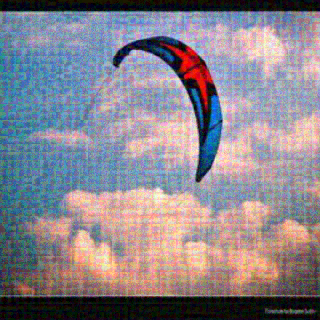,
            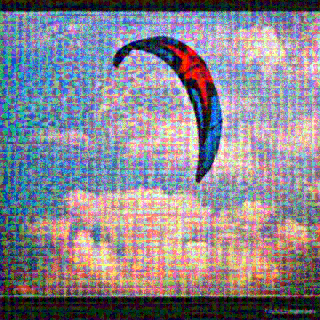,
            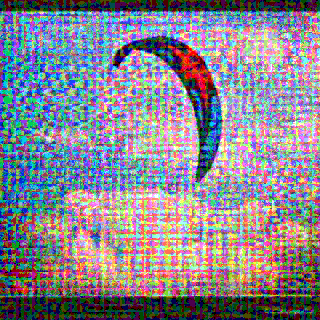
        }
        \place{}{}
        \place{ResNet18}{
            figure/apx/cam_rgbgolf_00.jpg,
            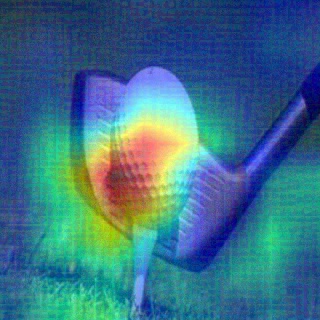,
            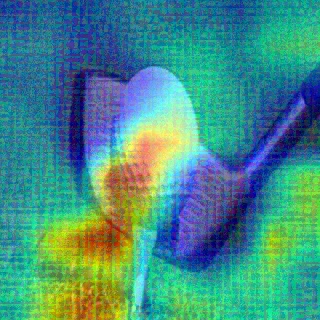,
            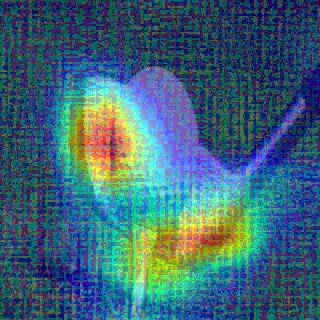,
            figure/apx/cam_rgbchute_00.jpg,
            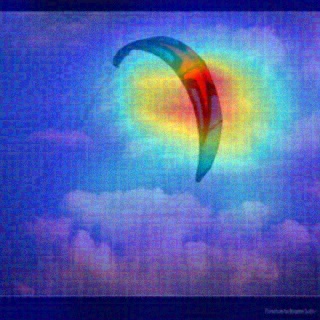,
            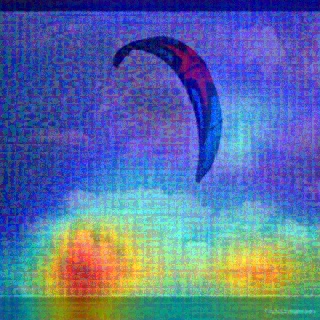,
            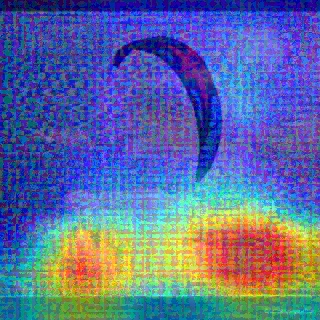
        }
        \place{EfficientNet}{
            figure/apx/cam_rgbefficient_golf00.jpg,
            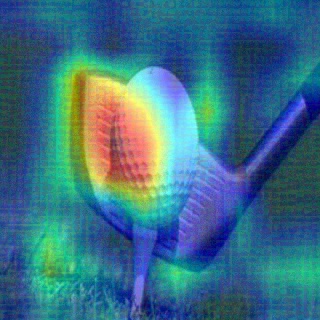,
            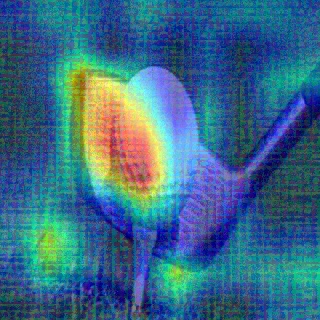,
            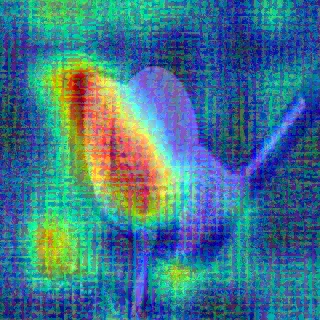,
            figure/apx/cam_rgbefficient_chute00.jpg,
            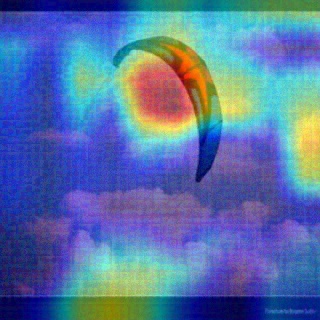,
            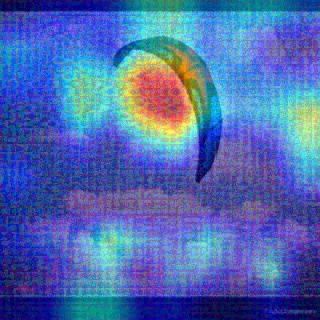,
            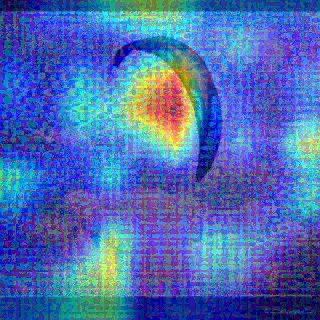,
        }
        \place{DenseNet}{
            figure/apx/cam_rgbdense_golf00.jpg,
            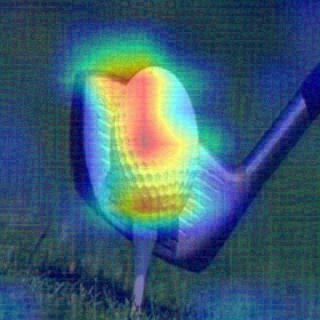,
            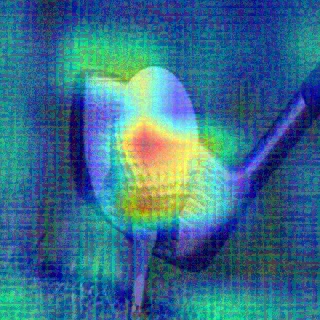,
            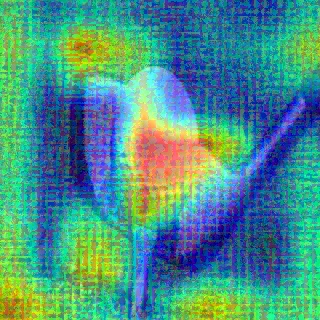,
            figure/apx/cam_rgbdense_chute00.jpg,
            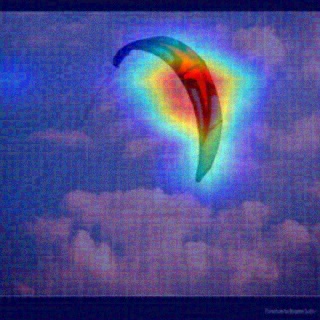,
            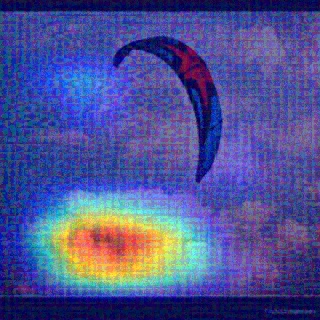,
            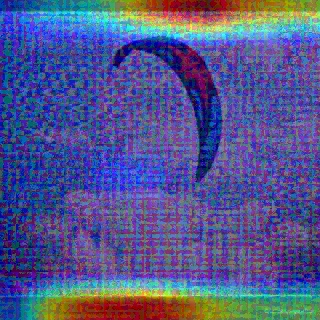,
        }
        \place{}{}
        \place{SF-ResNet18}{
            figure/apx/cam_fgolf_00.jpg,
            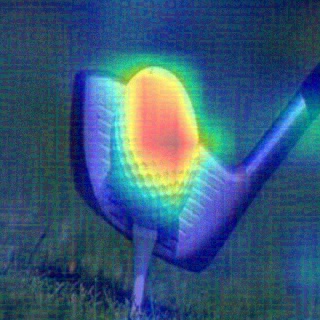,
            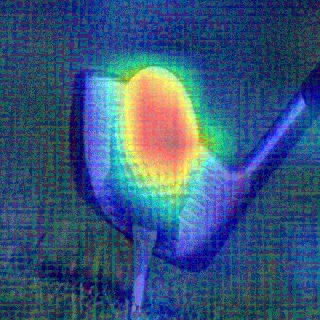,
            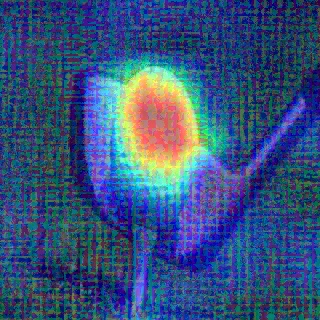,
            figure/apx/cam_fchute_00.jpg,
            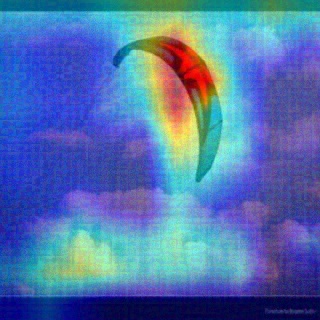,
            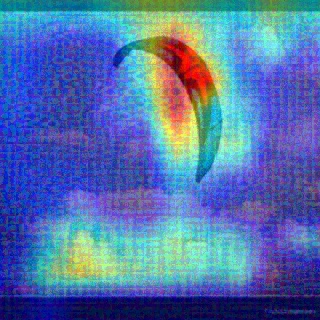,
            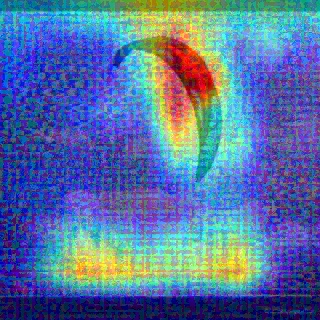
        }
        \place{SF-EfficientNet}{
            figure/apx/cam_fefficient_golf00.jpg,
            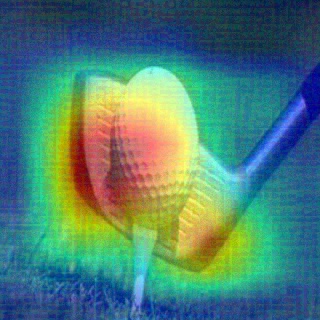,
            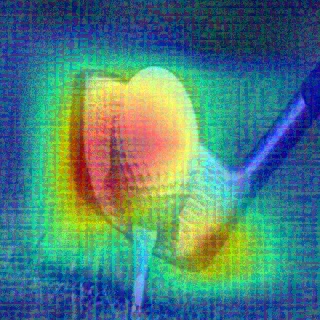,
            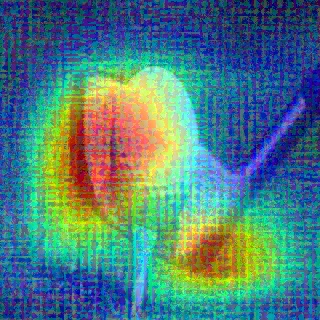,
            figure/apx/cam_fefficient_chute00.jpg,
            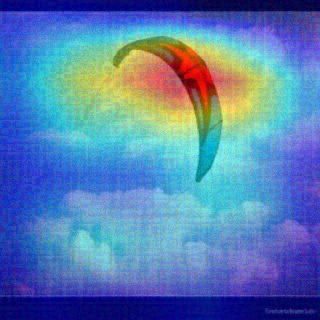,
            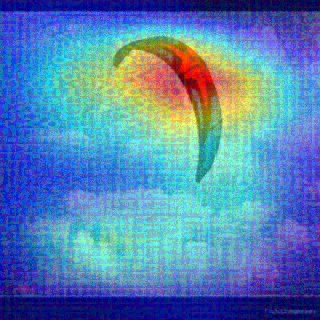,
            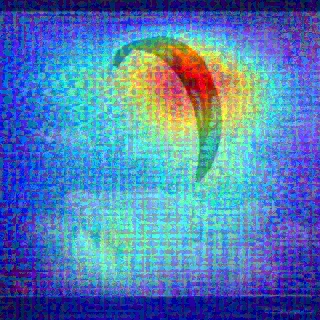
        }
        \place{SF-DenseNet}{
            figure/apx/cam_fdense_golf00.jpg,
            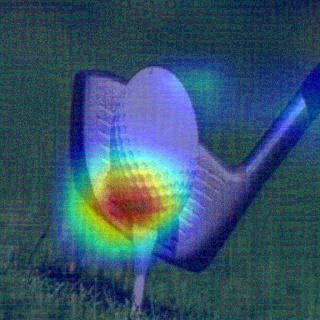,
            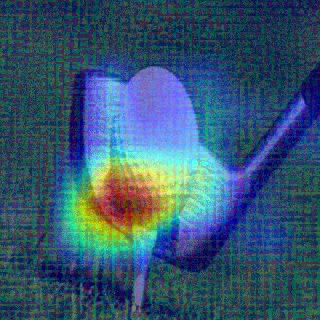,
            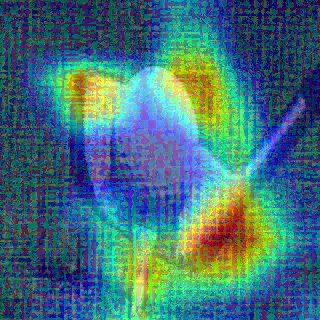,
            figure/apx/cam_fdense_chute00.jpg,
            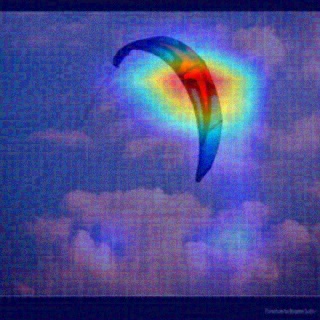,
            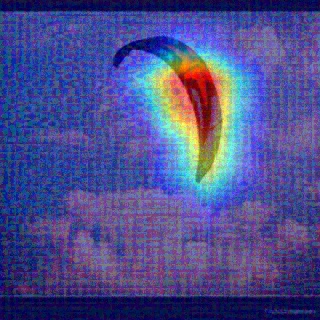,
            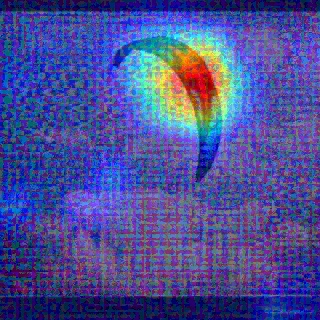,
        }
    \end{places}
    \caption{Grad-Cam visualizations of model attention on two adversarial examples generated from \textbf{SF-VGG11}. Each row shows the progression from $\epsilon=0$ (not attacked) to $\epsilon=0.1$, $0.2$ and $0.3$ from left to right.}
    \label{apxfig:grad_cam_sf}
\end{figure*}

Here, we provide the grad-cam visualizations of model attention for different CNN and SF-CNN models under transfer attacks from VGG11 in \cref{apxfig:grad_cam} and SF-VGG11 in \cref{apxfig:grad_cam_sf}. In both cases, SF-CNN models are consistently \emph{more focused}, i.e., the greater attention weight on the objects than their CNN counterparts. The visualizations support the claim that SF-CNNs learn a more robust set of features by operating in the spatial frequency domain.

\end{document}